\documentclass{article}

\usepackage{microtype}
\usepackage{graphicx}
\usepackage{subcaption}
\usepackage{booktabs} % for professional tables

\usepackage[table]{xcolor}
\usepackage{hyperref}

\NewDocumentCommand{\heng}
{ mO{} }{\textcolor{red}{\textsuperscript{\textit{Heng}}\textsf{\textbf{\small[#1]}}}}

\NewDocumentCommand{\jiateng}
{ mO{} }{\textcolor{orange}{\textsuperscript{\textit{jiateng}}\textsf{\textbf{\small[#1]}}}}

\NewDocumentCommand{\zhenhailong}
{ mO{} }{\textcolor{blue}{\textsuperscript{\textit{Zhenhailong}}\textsf{\textbf{\small[#1]}}}}

\NewDocumentCommand{\jeongh}
{ mO{} }{\textcolor{cyan}{\textsuperscript{\textit{Jeonghwan}}\textsf{\textbf{\small[#1]}}}}

\NewDocumentCommand{\aditi}
{ mO{} }{\textcolor{purple}{\textsuperscript{\textit{Aditi}}\textsf{\textbf{\small[#1]}}}}

\usepackage[preprint]{icml2026}

\usepackage{amsmath}
\usepackage{amssymb}
\usepackage{mathtools}
\usepackage{amsthm}

\usepackage{booktabs}
\usepackage{makecell}
\usepackage{multirow}
\usepackage{xcolor}
\usepackage{pifont}
\usepackage{comment}

\newcommand{\cmark}{\ding{51}}
\newcommand{\xmark}{\ding{55}}
\newcommand{\yes}{\textcolor{green!55!black}{\cmark}}
\newcommand{\no}{\textcolor{red!70!black}{\xmark}}

\usepackage[capitalize,noabbrev]{cleveref}

\theoremstyle{plain}

\theoremstyle{definition}

\theoremstyle{remark}

\usepackage[textsize=tiny]{todonotes}

\usepackage[most]{tcolorbox}

\newtcolorbox{PlannerPromptBox}{
  breakable,
  colback=white,
  colframe=blue!30,
  fonttitle=\bfseries,
  title=Planning Module Prompt,
}

\newtcolorbox{FeedbackPromptBox}{
  breakable,
  colback=white,
  colframe=orange!35,
  fonttitle=\bfseries,
  title=Feedback Module Prompt,
}

\newtcolorbox{FilePrepPromptBox}{
  breakable,
  colback=white,
  colframe=teal!30,
  fonttitle=\bfseries,
  title=File Preparation Planner Prompt,
}

\newtcolorbox{InitExplorePromptBox}{
  breakable,
  colback=white,
  colframe=blue!30,
  fonttitle=\bfseries,
  title=Planning Prompt: Initialize Exploration Targets,
}

\newtcolorbox{InitInteractPromptBox}{
  breakable,
  colback=white,
  colframe=blue!30,
  fonttitle=\bfseries,
  title=Planning Prompt: Initialize Shallow UI Interaction,
}

\newtcolorbox{StateClassifyPromptBox}{
  breakable,
  colback=white,
  colframe=orange!35,
  fonttitle=\bfseries,
  title=Feedback Prompt: State Classification,
}

\newtcolorbox{TerminalPromptBox}{
  breakable,
  colback=white,
  colframe=orange!35,
  fonttitle=\bfseries,
  title=Feedback Prompt: Terminal State Verification,
}

\newtcolorbox{PlanErrorPromptBox}{
  breakable,
  colback=white,
  colframe=orange!35,
  fonttitle=\bfseries,
  title=Feedback Prompt: Error Attribution (Planning),
}

\newtcolorbox{ActionErrorPromptBox}{
  breakable,
  colback=white,
  colframe=orange!35,
  fonttitle=\bfseries,
  title=Feedback Prompt: Error Attribution (Action),
}

\icmltitlerunning{OSExpert: Computer-Use Agents Learning Professional Skills via Exploration}

\begin{document}

\twocolumn[
  \icmltitle{OSExpert: Computer-Use Agents Learning Professional Skills via Exploration
  %Exploration-Based Skill Learning for Developing Professional Computer-Use Agents
  %OSExpert: Environment-Learned Computer-Use Agents with Professional Skills \heng{"Environment-learned" is grammatically awkward. maybe change to something like "Exploration based Learning for Professional Computer-Use Agent Development"}
  }



%\begin{comment}
\begin{center}
{\large\bfseries
Jiateng Liu$^{\heartsuit}$, Zhenhailong Wang$^{\heartsuit}$, Rushi Wang$^{\heartsuit}$, Bingxuan Li$^{\heartsuit}$, Jeonghwan Kim$^{\heartsuit}$,\\[0.35em]
Aditi Tiwari$^{\heartsuit}$, Pengfei Yu$^{\heartsuit}$, Denghui Zhang$^{\clubsuit}$, Heng Ji$^{\heartsuit}$
\par}

\vspace{0.5em}

{\large
$^{\heartsuit}$University of Illinois Urbana-Champaign  
$^{\clubsuit}$Stevens Institute of Technology
\par}

\vspace{0.25em}

{\normalsize
\texttt{\{jiateng5, hengji\}@illinois.edu}
\par}
\vspace{0.25em}

{Project Page: \url{https://oppugno-rushi.github.io/OSExpert}
\par}
\end{center}

\vspace{0.3in}

%\end{comment}
]



%\printAffiliationsAndNotice{}  % no special notice (required even if empty)

\begin{abstract}

General-purpose computer-use agents have shown impressive performance across diverse digital environments. However, our new benchmark, \textbf{OSExpert-Eval}, indicates they remain far less helpful than human experts. Although inference-time scaling enables adaptation, these agents complete complex tasks inefficiently with degraded performance, transfer poorly to unseen UIs, and struggle with fine-grained action sequences. To solve the problem, we introduce a GUI-based depth-first search (GUI-DFS) exploration algorithm to comprehensively explore and verify an environment’s unit functions. The agent then exploits compositionality between unit skills to self-construct a curriculum for composite tasks. To support fine-grained actions, we curate a database of action primitives for agents to discover during exploration; these are saved as a \textit{skill set} once the exploration is complete. We use the learned skills to improve the agent's performance and efficiency by (1) enriching agents with ready-to-use procedural knowledge, allowing them to plan only once for long trajectories and generate accurate actions, and (2) enabling them to end inference-time scaling earlier by realizing their boundary of capabilities. Extensive experiments show that our environment-learned agent takes a meaningful step toward expert-level computer use, achieving a $\sim$20\% performance gain on OSExpert-Eval and closing the efficiency gap to humans by $\sim$80\% \footnote{Data and code will be released at \url{https://github.com/Lumos-Jiateng/OSExpert}.}

\end{abstract}

\section{Introduction}
\begin{comment}
\heng{Are there any shared common hypotheses and innovations across these different solutions? Otherwise it seems a long list of solutions without a few coherent stories}
\heng{change "fine grained" to "fine-grained" throughout the paper}

\jiateng{These issues reflect different aspects of the non-professional nature of current computer-use agents. Much like human amateurs, existing agents attempt to solve a wide range of tasks using general prior knowledge, but lack the structured skills and reliability of true experts.}

\jiateng{We make agents behave more like experts by (1) learning expert knowledge through environment interaction, (2) supporting fine-grained execution with targeted hints and action primitives, and (3) making agents aware of the environment’s “world knowledge” and their own skill boundaries to improve reliability and efficiency.}
\end{comment}

%\heng{It will be good to elaborate what you mean by far from professional. categorize professional criteria into different dimensions, analyze reasons behind the limitations of existing methods, and which ideas you propose can address each limitation}

\begin{figure*}[t]
    \centering
    \includegraphics[width=1.0
    \textwidth]{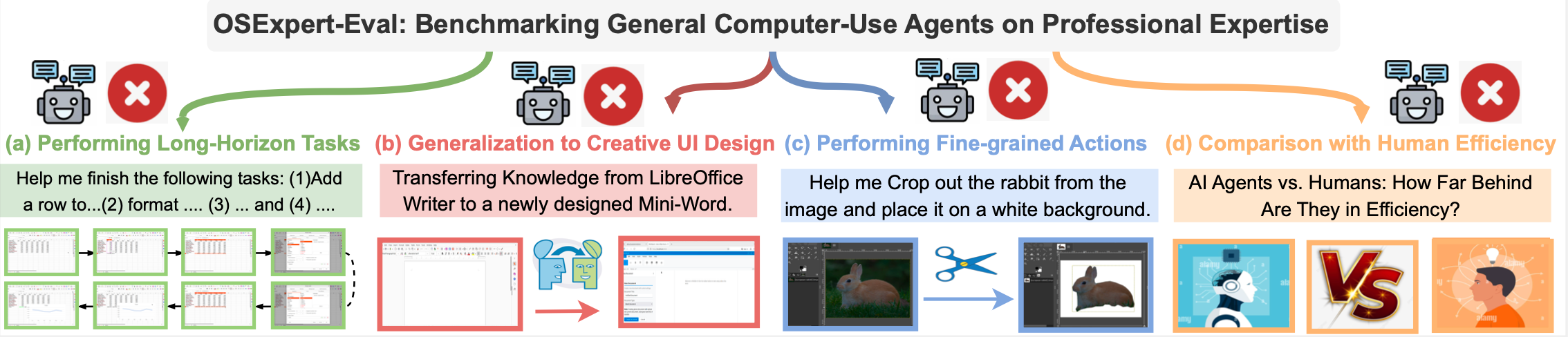}
    \caption{Our \textbf{OSExpert-Eval} shows that current computer-use agents remain far from expert-level: they struggle with long-horizon tasks, generalize poorly to unseen UI designs, lack fine-grained control over action sequences, and still fall well short of human expert efficiency. }
    \label{fig:motivation}
    \vspace{-12pt}
\end{figure*}

%General computer-use agents have recently demonstrated impressive abilities across a variety of digital environments: they can navigate file systems, assist with online shopping, browse the web, and operate everyday applications\citep{xie2024osworldbenchmarkingmultimodalagents,wang2024officebenchbenchmarkinglanguageagents,bonatti2024windowsagentarenaevaluating,pan2024webcanvasbenchmarkingwebagents,rawles2025androidworlddynamicbenchmarkingenvironment}. Powered by large multimodal models, these agents can interpret screens and execute grounded actions from natural-language instructions~\citep{agashe2024agentsopenagentic,wang2025opencua,hu2025osagentssurveymllmbased}; notably, Agent-S3 reoprts near–human-level performance on OSWorld~\citep{gonzalezpumariega2025unreasonableeffectivenessscalingagents,xie2024osworldbenchmarkingmultimodalagents}.

%\zhenhailong{the previous three sentences are a bit overly verbose, I would suggest to condense into one, or at most two}
%So, are today’s powerful computer-use agents good enough to solve real-world tasks?\zhenhailong{I would remove this question since it is not the core of this paper; and it sounds too big; we can highlight our key research question on the ``expert'' part; and it is a bit strange to directly jump in the new benchmark; would be good to first motivate why we need expert CUA}

General computer-use agents have demonstrated impressive abilities across diverse digital environments. Powered by large multimodal models, they are capable of navigating file systems, assisting with online shopping, browsing the web, and operating everyday applications by interpreting screens and executing grounded actions from natural-language instructions~\citep{xie2024osworldbenchmarkingmultimodalagents,wang2024officebenchbenchmarkinglanguageagents,bonatti2024windowsagentarenaevaluating,pan2024webcanvasbenchmarkingwebagents,rawles2025androidworlddynamicbenchmarkingenvironment,agashe2024agentsopenagentic,wang2025opencua,hu2025osagentssurveymllmbased}; Notably, Agent-S3~\citep{gonzalezpumariega2025unreasonableeffectivenessscalingagents} reports near–human-level performance on OSWorld~\citep{xie2024osworldbenchmarkingmultimodalagents}.

However, results on our new benchmark OSExpert-Eval indicate that current computer-use agents are not yet helpful enough for real-world workers operating professional software. As illustrated in Figure~\ref{fig:motivation}, OSExpert-Eval evaluates the competence of computer use agent along four dimensions: (1) completing complex, long-horizon tasks; (2) generalizing to unseen and creative UI designs; (3) executing fine-grained actions that require precise control; and (4) maintaining strong end-to-end performance relative to human experts who solve these tasks quickly and reliably. Across these dimensions, current agents consistently lag behind human experts: success rates decline sharply with increasing task complexity, dropping from near human-level to below 10\%. Generalization to unseen environments remains weak, and fine-grained execution frequently fails. Agents also rely on high-failure trial-and-error exploration, resulting in substantial inefficiency and 5–50× higher latency than human experts.

%In this work, we propose \textbf{OSExpert-Eval}, which suggests that current agentic frameworks have yet reach the proclaimed human-level performance. As illustrated in Figure~\ref{fig:motivation}, OSExpert-Eval evaluates the agent competence along the four dimensions: (1) the ability to complete complex, long-horizon tasks; (2) generalize to unseen and creative UI designs; (3) execute fine-grained actions that require precise control; and (4) retain overall performance relative to human experts who solve these tasks quickly and reliably. Across these dimensions, current agents consistently lag behind human experts; performance degrades as task complexity increases, generalization to unseen environments is weak, fine-grained actions often fail, and agents rely on slow trial-and-error exploration.

What went wrong? Our analysis of agent–environment interaction trajectories suggests a more fundamental limitation: it stems from how we train computer-use agents in the first place. Today’s foundation agents are largely trained on scaled human demonstrations across around $\sim$100 digital environments via behavior cloning and reinforcement learning~\cite{wang2025opencua,wang2025uitars2,ye2025mobileagentv3fundamentalagentsgui}. However, these agents often fail to acquire environment-specific procedural knowledge, and what they do learn transfers poorly as real-world interfaces and workflows evolve. Even in familiar environments, agents frequently solve long-horizon tasks through step-by-step planning and execution rather than directly obtaining reliable procedures, which introduces substantial computational overhead and amplifies cascading errors. In unfamiliar environments, the problem is more severe: agents struggle to identify the correct UI elements to interact with, and this difficulty compounds when tasks require fine-grained actions with precise control.

To address this problem, 
% and inspired by prior work advocating a focus on building agentic skills rather than constructing agents~\cite{zhang2025dontbuildagents}, 
%\zhenhailong{I removed the sentence in the middle; I don't think it aligns too well with our core idea and hurts the flow;}
we introduce OSExpert, an environment-learning paradigm designed to equip computer-use agents with expert knowledge and skills. 
%\zhenhailong{maybe: ... profesional knowledge and skills" (to echo the two problem aspects in previous paragraph)}
Instead of scaling human annotated data for new digital environments, OSExpert exposes agents directly to environments, enabling them to explore, autonomously acquire and compose their own skill sets without additional human effort for each environment. We further introduce three key technical innovations that show how this paradigm enables computer-use agents to acquire environment-specific knowledge, handle fine-grained action sequences, and achieve robust and efficient performance approaching the level of human experts.

First, we introduce a \textbf{GUI-DFS} exploration algorithm to systematically discover the environment’s unit functions. The exploration is carried out by three coordinated modules: (i) a \textit{planning module} that generates exploration plans, (ii) an \textit{action module} that executes UI actions and interacts with the environment, and (iii) a \textit{feedback module} that evaluates the current exploration state and provides state-dependent feedback. We detail the design of these modules and their interactions in Section~\ref{sec:methods}. Based on the discovered unit functions, the agent then self-proposes a learning curriculum by composing unit skills into higher-level procedures, which allows it to acquire procedural knowledge beyond the limits of common user-query distribution. The discovered unit functions and the validated composite procedures are both consolidated into the agent’s skill set. %\zhenhailong{too long single paragraph; cut new line}

%Second, to improve inference efficiency, we go beyond attaching reference trajectories for step-by-step imitation and test-time scaling. Instead, we (1) apply LoRA to fine-tune a lightweight language model to generate the entire planning sequence in a single forward pass, removing the latency and error accumulation associated with step-wise planning, In multi-environment settings, we only store and load the \textbf{environment-specific LoRA weights} alongside the skill set, providing a compact mechanism to retain both world knowledge and procedural knowledge. (2) we ask the agent to check its knowledge boundary before assisting with users. By verifying failed and unsolved exploration traces, the agent realizes its limitations and stop early to avoid wasting user time. 

Second, to improve inference efficiency, we go beyond step-by-step imitation via reference trajectories and blind test-time scaling. Instead, we: (1) apply LoRA to fine-tune a lightweight language model that generates an entire plan in a single forward pass, reducing latency and avoiding the error accumulation of iterative planning. %In multi-environment settings, we store and load only the \textbf{environment-specific LoRA weights} alongside the skill set, providing a compact mechanism to retain both world knowledge and procedural knowledge. 
(2) introduce a skill-boundary check that predicts when additional test-time scaling would be unproductive; by validating failed and unsolved exploration traces, the agent can recognize low-success regimes and terminate early, avoiding the wasted latency of blind rollout-based exploration.

%\jeongh{Is there a comparison experiment between step-wise planning vs. LoRA equipped LM on these two criteria?}.\jiateng{not yet, i will add those in rebuttal, including experiments for baseline + fine-grained actions}

Third, to construct skills for accurate fine-grained manipulation and control, we provide a database of fine-grained action descriptions, action-sequence primitives, and grounding modules for constructing task-specific solutions. When the feedback module indicates a need for fine-grained control, the agent selects a promising primitive, executes it with grounding-based verification, and consolidates successful solutions into the skill set for reuse at inference time.

We deploy our learning framework across multiple digital environments and evaluate it on OSExpert-Eval. The results consistently validate our design and represent a substantial step toward expert-level performance without requiring any additional human effort. In particular, while existing computer-use agents peak at roughly $\sim$10\% success on long-horizon tasks, our agent improves success rate to around 30\%. Moreover, our agent transfers reliably to unseen UI environments and executes fine-grained actions with high fidelity with the obtained skill set, achieving around $\sim$20\% success gain and improving efficiency by roughly $\sim$80\% relative to the most efficient existing agent.

\section{OSExpert-Eval: Benchmarking General CUAs on Professional Expertise}
\label{sec:data}

\begin{table*}[!t]
\centering
\caption{Comparison of web-based agents and computer-use agent benchmarks. Most mainstream benchmarks do not explicitly evaluate composite skills, generalization to creative UI designs, fine-grained low-level interaction sequences, or latency \& efficiency evaluation.}
\label{tab:bench}

\label{tab:benchmark_compare_osexpert}
\setlength{\tabcolsep}{6pt}
\renewcommand{\arraystretch}{1.15}
\resizebox{\textwidth}{!}{%
\begin{tabular}{lcccc}
\toprule
\textbf{Real-World Benchmark /\ Evaluation Dimensions} &
\makecell{\textbf{Long-Horizon}\\\textbf{Composite Skills}} &
\makecell{\textbf{Creative UI}\\\textbf{Generalization}} &
\makecell{\textbf{Fine-Grained}\\\textbf{UI Interaction}} &
\makecell{\textbf{Latency}\\\textbf{Evalaution}} \\
\midrule
Mind2Web~\cite{deng2023mind2web}             & \yes & \no & \no & \no \\
%WorkArena~\cite{drouin2024workarena}        & \yes & \no  & \no & \no \\
BEARCUBS~\cite{song2025bearcubs}            & \yes & \no & \no & \no \\
\midrule
OSWorld~\cite{xie2024osworld}                & \no & \no & \no & \no \\
WinAgentArena~\cite{bonatti2024windowsagentarena}  & \yes & \no & \no & \no \\
%AndroidWorld~\cite{rawles2024androidworld}   & \no & \no  & \no & \no \\
macOSWorld~\cite{yang2025macosworldmultilingualinteractivebenchmark}   & \yes & \no & \no & \no \\
OSWorld-Human~\cite{abhyankar2025osworldhuman}  & \no & \no & \no & \yes \\
OSUniverse~\cite{davydova2025osuniverse}     & \yes & \no  & \yes & \no \\
\midrule
\textbf{Ours (OSExpert-Eval)}                     & \yes & \yes & \yes & \yes \\
\bottomrule
\end{tabular}}
\end{table*}

While recent work reports human-level results on OSWorld~\cite{xie2024osworld}, the benchmark itself is limited to relatively low task complexity and a narrow set of evaluation criteria that do not full reflect human-level capabilities. As a testatment, current computer-use agents still fall short of human experts in real application scenario, with slower interaction and inconsistent response quality. To quantify this gap, we introduce \textbf{OSExpert-Eval}, a new benchmark that evaluates agent expertise across four dimensions, as shown in Figure~\ref{fig:motivation}. OSExpert-Eval measures: 

\textbf{(1) Long-horizon compositional workflows}, which require chaining multiple unit functions to achieve higher-level objectives beyond the short, single-function tasks in OSWorld~\cite{xie2024osworld}. We use the same environments, including GIMP\footnote{GIMP is an open-source image editing tool on Linux.} and LibreOffice, but increase only task compositionality and complexity to isolate and quantify the performance gap induced by OSWorld’s simplified tasks.

\textbf{(2) Generalization to unseen and creative user interfaces}, which requires robust operation under customized layouts, icon-centric toolbars, and non-standard interaction patterns; we include the Tableau environment, a data visualization tool for data analysts that is rarely represented in the training distribution of current fundamental GUI agents and contains multiple out-of-distribution interaction patterns, and we also include our self-designed MiniWord editor, designed with creative UI layouts and vivid icons to minimize memorized interface conventions of similar environments. %such as Microsoft Word and Google Docs.
%despite its simplicity and shared functionalities with Microsoft Word, MiniWord is 

\textbf{(3) Fine-grained low-level actions}, which require precise spatial control and accurate localization of interaction targets, where minor deviations can cause task failure. This stream includes tasks such as selecting target text, drag-and-drop, tracing object boundaries in images, and performing rotation or translation with specified angles or sizes.

\textbf{(4) Execution efficiency}, which measures the practical cost of attempting a task in terms of elapsed time, regardless of success or failure. We report human expert efficiency based on operators who are familiar with the tasks and practiced them prior to evaluation.

In total, OSExpert-Eval contains 113 evaluation tasks. Table~\ref{tab:bench} contrasts OSExpert-Eval with prior benchmarks and summarizes the key distinctions. Additional benchmark details and case examples are provided in Appendix~\ref{app:data detail}.

\section{Environment-Learned CUAs}
\label{sec:methods}

To equip computer-use agents with professional skills in a scalable and generalizable way, OSExpert directly exposes the agent to target digital environments and learns verifiable skills through interaction. As shown in Figure~\ref{fig:comprehensive} and Figure~\ref{fig:fine-grained}, OSExpert improves both performance and efficiency through the following highlights: (1) a bottom-up exploration paradigm that comprehensively discovers unit-level functions for expert-level environment specific knowledge; (2) a lightweight LoRA-tuned planner that generates a complete plan in a single forward pass to avoid step-wise plan latency, together with a skill boundary check that avoids blind test-time scaling for expert efficiency; and (3) predefined action primitives and grounding modules that are verified during exploration and consolidated into the skill set for expert manipulation and control.

\subsection{Bottom-Up Self-Exploration for Expert Knowledge}
\label{sec:3.1}
\begin{figure*}[t]
    \centering
    \includegraphics[width=1.0
    \textwidth]{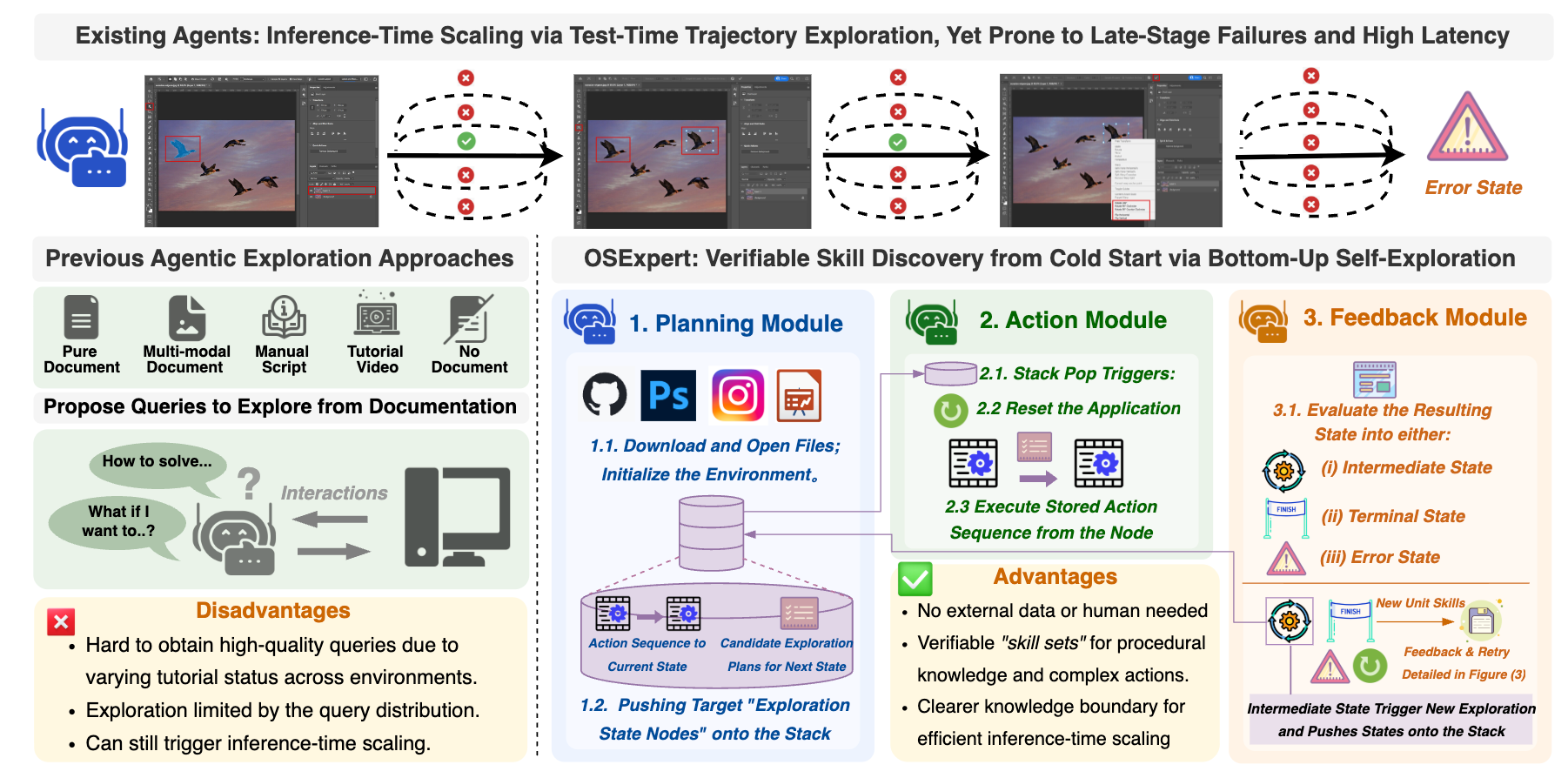}
    \caption{\textbf{Up:} Current general-purpose computer-use agents rely on inference-time scaling, yet remain prone to failures and high latency. \textbf{Left:} Prior approaches explore digital environments using human-curated queries or tutorial-derived queries, which are often unavailable or difficult to obtain for arbitrary environments. \textbf{Right:} Our framework does not require external data or human effort for exploration queries and more comprehensively discover the unit functions of the digital environment, and benefits both performance and efficiency. We introduce how we handle the fine-grained actions during the exploration and how we  organize the learned skill set in Figure~\ref{fig:fine-grained}.}
    \label{fig:comprehensive}
    \vspace{-12pt}
\end{figure*}

%\zhenhailong{currently this section is very hard to read; can we further use some paragraph titles to summarize each procedure? for example, Initialization; Exploration; Outcome of the Exploration; Inference}
As shown in Algorithm~\ref{alg:gui_DFS_reverse}, our core design is a GUI-DFS procedure that comprehensively discovers unit-level functionalities of a digital environment first, and later organizes and consolidates them into a skill set for robust and efficient inference. As illustrated in Figure~\ref{fig:comprehensive}, this approach reduces reliance on inference-time scaling and enables exploration without additional human effort, such as curating queries from tutorials or providing environment-specific use cases.

\textbf{Initialization:} Before exploration, the agent initializes three coordinated modules: a planning module, an action module, and a feedback module. As shown in Figure~\ref{fig:comprehensive}, the planning module forms lightweight assumptions about the application’s data types and prepares downloaded or self generated template inputs. The action module then performs shallow interactions and UI analysis to infer the hierarchical structure of visible interface elements on the initial screen. Based on this structure, the agent enumerates top layer UI elements and selects an initial subset as targets to start the DFS based exploration. Each selected target is pushed onto the DFS stack as an exploration state node containing the current screenshot, the plan and action sequence to reach the state, and the next step plan conditioned on the state.
%We construct the target agent’s skill set by acquiring procedural knowledge through bottom-up self-exploration of unit functions in the target digital environment. As illustrated in Algorithm~\ref{alg:gui_DFS_reverse}, we design a GUI-DFS algorithm with reverse traversal to discover unit functionalities as comprehensively as possible. 

\textbf{Exploration with GUI-DFS:} As shown in Figure~\ref{fig:comprehensive}, the algorithm follows a DFS stack discipline and iteratively pops the top exploration state node in a last in first out manner. To recover the node state, the environment is restarted and the stored action sequence is replayed. The action module then predicts and executes actions to satisfy the node’s next step plan. After execution, the feedback module evaluates the new state with respect to the exploration history and classifies it as an intermediate state, a terminal state, or an error state due to an invalid plan or action. For intermediate states, the planning module proposes follow up plans, updates the action sequence to reach each new state, and pushes the resulting state nodes onto the stack. For terminal states, the validated plan and action sequence are condensed into a unit function skill and added to the skill set with a short description of when and how to use it. For error states, the feedback module provides targeted critiques of the plan or action, and the agent revises and retries for a bounded number of attempts. If failures persist, the error case is summarized and recorded in the skill set. Fine-grained action handling from error states is described in Section~\ref{sec:3.3}.

\begin{figure*}[t]
    \centering
    \includegraphics[width=1.0
    \textwidth]{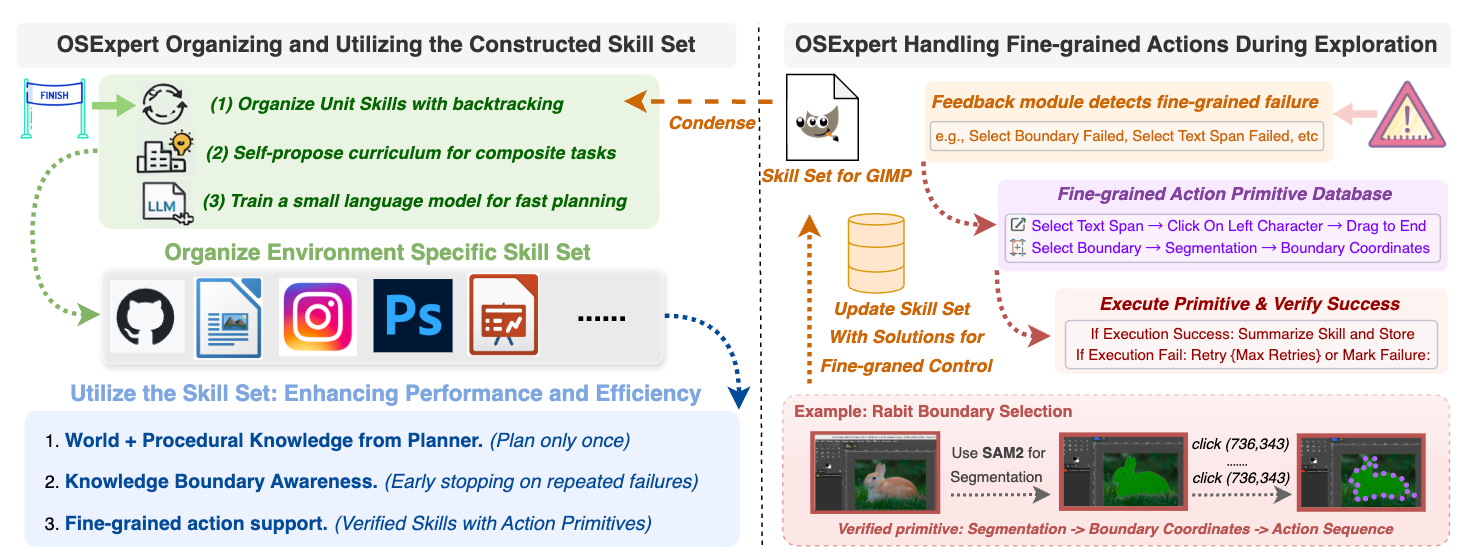}
    \caption{\textbf{Left:} How our framework organizes and utilize the self-constructed skill set for robust and efficient inference. The unit functions are obtained from the terminal states as shown in Figure~\ref{fig:comprehensive}. \textbf{Right:} How we handled potential fine-grained actions during exploration stage. The fine-grained action handling is usually triggered by an error state in the exploration, as shown in Figure~\ref{fig:comprehensive}. The selected primitive fine-grained action template will be added to the skill set for solving future queries if verified helpful.}
    \label{fig:fine-grained}
    \vspace{-12pt}
\end{figure*}

\textbf{Organize and Extend the Skill Set:} As shown in Figure~\ref{fig:fine-grained} (left), the exploration stage produces a set of unit function skills by backtracking to each initial UI entry and systematically exploring its available functions. Each successful skill stores the planning sequence and action sequence required to complete the corresponding unit operation, while failed attempts are recorded as exploration failures. We further extend the skill set by prompting the agent to propose a curriculum of composite tasks that leverage natural compositions of unit functions, and we add the resulting composite skills to the unit skill set. To organize the skills, the agent summarizes what each skill can do, and we train a small language model to map each summary to its corresponding planning sequence, improving efficiency as discussed in Section~\ref{sec:3.2}. Skills associated with error states can be improved by pushing the corresponding nodes back onto the stack for additional exploration with stronger agents, or by manual curation from human experts. In our experiments, we run exploration only once and use no human curation.

\begin{algorithm}[t]
\caption{GUI-DFS Algorithm
}
\label{alg:gui_DFS_reverse}

{\small
\textbf{Definitions.}
Digital environment $E$; Environment State $S_i$; Exploration Plans $\Pi$; Action Sequence $\alpha$;
Exploration State Node
$n \triangleq (\Pi,\alpha)$; Exploration Outcome $T \in \{\mathtt{Continue},\mathtt{Final},\mathtt{Error}\}$.

\par\textbf{Inputs.}
Planner Module $P$; Action Module $A$; Feedback Module $F$; Max Retries $R$.

\textbf{Output.}
Unit Skill Set $\mathcal{K}$ for Environment $E$.
\par}

\begin{algorithmic}[1]
\STATE $\mathcal{Q} \leftarrow \emptyset,\ \mathcal{K} \leftarrow \emptyset$, $(S_0) \leftarrow \texttt{Reset}(E)$
\STATE $\mathcal{Q} \leftarrow \texttt{PushAll}(\mathcal{Q},\ P(E,S_0))$ 
\WHILE{$\mathcal{Q} \neq \emptyset$}
  \STATE $(\Pi,\alpha) \leftarrow \texttt{Pop}(\mathcal{Q})$; $(S_0) \leftarrow \texttt{Reset}(E)$
  \STATE $S_i \leftarrow \texttt{Execute}(E,\alpha)$; $r \leftarrow 0$
  \WHILE{$r < R$}
    \STATE $\alpha' \leftarrow A(\Pi,\alpha,S_i)$; $S_{i+1} \leftarrow \texttt{Execute}(E,S_i,\alpha')$
    \STATE $(T,feedback)\leftarrow F(S_i,S_{i+1},\Pi,\alpha')$
    \IF{$T=\mathtt{Continue}$}
      \STATE $\mathcal{K} \leftarrow \mathcal{K} \cup \{(\Pi,\ \alpha \oplus \alpha')\}$
      \STATE \textbf{for} each $\Pi^{new} \in P(E,S_{i+1},\Pi,\alpha \oplus \alpha')$ \textbf{do}
      \STATE \hspace{1em}$\mathcal{Q} \leftarrow \texttt{Push}\!\left(\mathcal{Q},\ (\Pi \oplus \Pi^{new},\ (\alpha \oplus \alpha') )\right)$
      \STATE \textbf{end for}
      \STATE \textbf{break}
    \ELSIF{$T=\mathtt{Final}$}
      \STATE $\mathcal{K} \leftarrow \mathcal{K} \cup \{(\Pi,\ \alpha \oplus \alpha')\}$
      \STATE \textbf{break}
    \ELSE 
      \STATE $\mathcal{Q} \leftarrow \texttt{Push}\!\left(\mathcal{Q},\ (\Pi \oplus feedback,\ \alpha \oplus \alpha')\right)$
      \STATE $r \leftarrow r + 1$
      \STATE \textbf{break}
    \ENDIF
  \ENDWHILE
  \IF{$r \ge R$} 
  \STATE $\mathcal{K} \leftarrow \mathcal{K} \cup \{(\Pi,\ \alpha)\}$
  \ENDIF
\ENDWHILE
\STATE \textbf{return} $\mathcal{K}$
\end{algorithmic}
\end{algorithm}

\subsection{Fast Planner and Skill Check for Expert Efficiency}
\label{sec:3.2}

%\zhenhailong{we should add a pointer in Sec 3.1 to this section showing that this is build upon the exploration outcomes}

%\zhenhailong{I feel the use of the term skill is a bit confusing; as we use procedural knowledge in previous sections; can we use knowledge instead? and maybe use skill to refer to the sec 3.3 fine-grained actions as skill?}

We adopt two approaches to reduce latency when using the skill set obtained in Section~\ref{sec:3.1}, each addressing a distinct bottleneck in existing computer use agents. First, agents often require many more interaction steps than humans and incur high overhead from step wise planning without procedural knowledge. Second, agents spend substantial time on inference time scaling for difficult tasks that still end in failure. Accordingly, we use a fast planner to generate a complete plan in a single forward pass, reducing step wise planning and extra interactions, and we apply a skill boundary check to early stop on tasks deemed infeasible by the skill set, avoiding wasted inference time exploration.

\textbf{Single Pass Fast Planner:} Prior work~\cite{abhyankar2025osworldhuman} shows that current agents take about 1.4× to 2.7× more interaction steps than humans, and step-wise planning accounts for nearly 50\% of total latency on OSWorld~\cite{xie2024osworld}. Our learned skill set provides procedural knowledge that support direct plan induction and reduce mid execution hesitations. We use the small language model trained on task planning pairs in Section~\ref{sec:3.1} to generate a full plan in a single forward pass. The planner is LoRA tuned, and we store only the fine tuned weights for each digital environment, enabling low overhead deployment. Note that the fast planner reduces planning overhead but does not remove per step perception and action selection: the action module still inspects the screen at each step to produce the next action. When execution fails, the system falls back to the general planning stack, where relevant skills are injected into the agent context for test time scaling.

%A major source of inference time latency in computer use agents is the large number of step wise planning decisions required during execution. Prior work shows that current agents take about 1.4× to 2.7× more interaction steps than humans, and that step wise planning accounts for nearly 50\% of total latency on OSWorld~\cite{abhyankar2025osworldhuman,xie2024osworld}. This inefficiency largely reflects limited world and procedural knowledge: without strong priors, agents must repeatedly inspect intermediate screens to infer the next step, leading to excessive deliberation and compounding latency.

\textbf{Skill Boundary Check for Early Stopping:} As shown in Figure~\ref{fig:comprehensive} (top), current computer use agents are typically unaware of their skill boundaries. When test time scaling has remaining iterations, they often continue attempting infeasible solutions, which leads to failure with substantial latency. Our learned skill set provides an explicit notion of capability. In particular, failed entries record unit functions that the agent repeatedly attempted during exploration but could not complete within bounded retries, suggesting low likelihood of success even with additional trials. Therefore, if an input query maps to a skill marked as failed, the agent stops early and reports an error, avoiding prolonged futile attempts during inference.

\subsection{Skill Construction for Expert Fine-grained Control}
\label{sec:3.3}

We observe that many GUI manipulation tasks require precise visual grounding and accurate low level action generation. However, training data for such fine-grained behaviors remains scarce, causing current computer use agents to struggle with operations such as selecting an exact text span, dragging with pixel level precision, rotating objects by a specified angle, or tracing object boundaries in images. Without dedicated handling, the skill set constructed in Section~\ref{sec:3.1} can accumulate many failure states.

To address this issue, we identify recurring fine-grained action primitives and organize them into a database with concise descriptions. As indicated in Figure~\ref{fig:fine-grained} (right), we use this dataset to support skill construction during exploration. Specifically, when a fine-grained action is required, the feedback module typically triggers an error state and, based on the current exploration context, searches the database for suitable action primitives that may resolve the situation. For the GIMP scissor select tool usage example shown in figure, the matched primitive specifies calling an external segmentation tool, extracting boundary coordinates, and clicking a sequence of points along the object contour. We add the procedure to the skill set only if it succeeds under verification, together with a brief condition describing when to invoke the primitive. This process requires minimal human effort beyond defining common fine-grained patterns and providing feasible solutions to the agent. It substantially reduces annotation cost compared to approaches that rely on large scale manual labeling and supervised training for fine-grained control~\cite{hu2025showuipiflowbasedgenerativemodels}.

\section{Experiments}

\subsection{Experimental Settings}

\paragraph{Environment and Benchmark}
We evaluate current computer-use agents and our environment learning framework on OSExpert-Eval, which consists of six interactive GUI environments. These include LibreOffice Writer, LibreOffice Calc, LibreOffice Impress, GIMP, web-based Tableau Public, and \textsc{MiniWord}, a lightweight text-editing environment developed in-house.
These environments  are detailed in Section~\ref{sec:data} and Appendix~\ref{app:data detail}.

\paragraph{Model Configuration}
We use different model configurations for exploration and inference. During exploration, we use either (1) GPT-5 as the high-level planner and feedback module, and UI-TARS-1.5-7B~\cite{qin2025uitarspioneeringautomatedgui} as the action module, or (2) Qwen-3-VL-8B for all modules. We set the maximum retry budget to $R=4$. We further include auxiliary grounding tools for fine-grained perception and verification, using Qwen-3-VL~\cite{bai2025qwen3vltechnicalreport} for grounding verification and SAM2~\cite{ravi2024sam2segmentimages} for segmentation. After environment learning, we switch to inference using the constructed skill set and fast planning cache. We use Qwen-3-VL-8B as the base inference agent and Qwen-3-4B as the fast planner. Following Section~\ref{sec:3.1} and Section~\ref{sec:3.2}, we first execute the plan generated by the fast planner for an input query, and only fall back to test-time scaling with relevant skills injected into the model context if execution fails. All experiments use the automatically constructed skill set without manual correction. For baselines, we include representative computer-use agents covering (i) specialized systems (e.g., OpenCUA), (ii) general-purpose multimodal LLMs (e.g., Qwen-3-VL-8B), and (iii) agentic frameworks (e.g., Agent-S ). These methods typically rely on test-time scaling to adapt to new environments.

\paragraph{Evaluation Protocol}
To ensure fair comparison across methods, we enforce a unified interaction budget for all agents. Most OSExpert-Eval tasks require fewer than 10 interaction steps for a human expert, and the longest-horizon task takes 14 steps; we therefore cap each episode at 30 interaction steps for all methods. We always set base model temperature to 1.0. We use the official API for OpenAI moels, and serve open-source models via vLLM~\cite{kwon2023efficient}. We report average task success rate and average task completion time for three independent runs over the benchmark in Tables~\ref{tab:method_compare_performance} and~\ref{tab:method_compare_efficiency}, and we report results separately for each environment. More detailed results, including computational budget estimates, and results on general computer-use benchmarks, are provided in Appendix~\ref{app:Details}.

\begin{table*}[!t]
\centering
\caption{Performance comparison on \textbf{OSExpert-Eval} using \textbf{average success rate ($\uparrow$)}. Our agent acquires expert-level skills for long-horizon tasks, generalizes effectively to unseen UIs, and executes fine-grained actions reliably.}
\label{tab:method_compare_performance}
\setlength{\tabcolsep}{8pt}
\renewcommand{\arraystretch}{1.2}
\resizebox{\textwidth}{!}{%
\begin{tabular}{
llcccccc
}
\toprule
\multirow{2}{*}{{\large\textbf{General CUAs}}} &
\multirow{2}{*}{{\large\textbf{Agent Type}}} &
\multicolumn{2}{c}{\textbf{Long-Horizon Composite Skills}} &
\multicolumn{2}{c}{\textbf{Unseen UI Generalization}} &
\multicolumn{2}{c}{\textbf{Fine-Grained Action Execution}} \\
\cmidrule(lr){3-4}\cmidrule(lr){5-6}\cmidrule(lr){7-8}
& &
\textbf{GIMP} & \textbf{LibreOffice} &
\textbf{Tableau} & \textbf{MiniWord} &
\textbf{GIMP} & \textbf{LibreOffice} \\
\midrule
OpenCUA-7B~\cite{wang2025opencua}
& Specialized Model
& 0.00 $\pm$ 0.00  & 0.00 $\pm$ 0.00 & 0.02 $\pm$ 0.03 & 0.00 $\pm$ 0.00  & 0.07 $\pm$ 0.00  & 0.05 $\pm$ 0.00  \\

Computer-Use-Preview~\cite{openai2025computerusepreview}
& Specialized Model
& 0.00 $\pm$ 0.00 & 0.04 $\pm$ 0.00  & 0.00 $\pm$ 0.00  & 0.00 $\pm$ 0.00 & 0.00 $\pm$ 0.00 & 0.00 $\pm$ 0.00 \\

Qwen-3-VL-8B~\cite{bai2025qwen3vltechnicalreport}
& General Model
& 0.00 $\pm$ 0.00  & 0.00 $\pm$ 0.00   & 0.00 $\pm$ 0.00 & 0.00 $\pm$ 0.00 & 0.07 $\pm$ 0.00 & \underline{0.10 $\pm$ 0.00}  \\

%Agent-S2.5 w/ o3~\cite{agashe2025agents2}
%& Agentic Framework
%& 0.00 & 0.00 & 0.00 & 0.00 & 0.00 & 0.00 \\

CoAct-1~\cite{song2025coact1}
& Agentic Framework
& 0.00 $\pm$ 0.00 & 0.08 $\pm$ 0.00 & 0.05 $\pm$ 0.00  & 0.06 $\pm$ 0.02 & 0.00 $\pm$ 0.00 & 0.05 $\pm$ 0.00\\

Agent-S3 w/ GPT-5~\cite{gonzalezpumariega2025unreasonableeffectivenessscalingagents}
& Agentic Framework
& \underline{0.06 $\pm$ 0.10} & 0.08 $\pm$ 0.00 & 0.03 $\pm$ 0.03 & 0.02 $\pm$ 0.02 & 0.00 $\pm$ 0.00 & \underline{0.10 $\pm$ 0.00} \\

\midrule
%\rowcolor{gray!20}
%\multicolumn{8}{c}{\textbf{Complete OSExpert Framework}} \\
\textbf{OSExpert (Explore w/ Qwen-3-VL-8B, wo/ Fine Skills)}
& Agentic Framework
& \textbf{0.33 $\pm$ 0.00} & \underline{0.29 $\pm$ 0.00}  & \underline{0.23 $\pm$ 0.03} & \underline{0.30 $\pm$ 0.00} & \underline{0.14 $\pm$ 0.00} & \underline{0.10 $\pm$ 0.00} \\
\textbf{OSExpert (Explore w/ Qwen-3-VL-8B)}
& Agentic Framework
& \textbf{0.33 $\pm$ 0.00} & \underline{0.29 $\pm$ 0.00}  & \textbf{0.25 $\pm$ 0.00} & \textbf{0.37 $\pm$ 0.00} & \textbf{0.28 $\pm$ 0.00} & \textbf{0.26 $\pm$ 0.00} \\
\textbf{OSExpert (Explore w/ GPT-5 \& UI-TARS)}
& Agentic Framework
& \textbf{0.33 $\pm$ 0.00} & \textbf{0.31 $\pm$ 0.02}  & \textbf{0.25 $\pm$ 0.00} & \textbf{0.37 $\pm$ 0.00} & \textbf{0.28 $\pm$ 0.00} & \textbf{0.26 $\pm$ 0.00} \\
\bottomrule
\end{tabular}}
\end{table*}

\begin{table*}[!t]
\centering
\caption{Efficiency comparison on \textbf{OSExpert-Eval}, measured by \textbf{average execution time (seconds, $\downarrow$)}. Our agent substantially narrows the efficiency gap between computer-use agents and human experts.}
\label{tab:method_compare_efficiency}
\setlength{\tabcolsep}{8pt}
\renewcommand{\arraystretch}{1.2}
\resizebox{\textwidth}{!}{%
\begin{tabular}{
llcccccc
}
\toprule
\multirow{2}{*}{{\large\textbf{General CUAs}}} &
\multirow{2}{*}{{\large\textbf{Agent Type}}} &
\multicolumn{2}{c}{\textbf{Long-Horizon Composite Skills}} &
\multicolumn{2}{c}{\textbf{Unseen UI Generalization}} &
\multicolumn{2}{c}{\textbf{Fine-Grained Action Execution}} \\
\cmidrule(lr){3-4}\cmidrule(lr){5-6}\cmidrule(lr){7-8}
& &
\textbf{GIMP} & \textbf{LibreOffice} &
\textbf{Tableau} & \textbf{MiniWord} &
\textbf{GIMP} & \textbf{LibreOffice} \\
\midrule
OpenCUA-7B~\cite{wang2025opencua}
& Specialized Model
& 169 $\pm$ 5 & 193 $\pm$ 11 & 216 $\pm$ 6 & 235 $\pm$ 2 & 162 $\pm$ 7 & 173 $\pm$ 3 \\

Computer-Use-Preview~\cite{openai2025computerusepreview}
& Specialized Model
& 252 $\pm$ 15 & 240 $\pm$ 9 & 361 $\pm$ 11 & 296 $\pm$ 2 & 256 $\pm$ 14  & 282 $\pm$ 7 \\

Qwen-3-VL-8B~\cite{bai2025qwen3vltechnicalreport}
& General Model
& 139 $\pm$ 2 & 136 $\pm$ 4 & 135 $\pm$ 4 & 153 $\pm$ 1 & 83 $\pm$ 5 & 138 $\pm$ 3 \\

%Agent-S2.5 w/ o3~\cite{agashe2025agents2}
%& Agentic Framework
%& 834 & 850 & 513 & 487 & 549 & 462 \\

CoAct-1~\cite{song2025coact1}
& Agentic Framework
& 1260 $\pm$ 58 & 1104 $\pm$ 73 & 924 $\pm$ 31 & 857 $\pm$ 61 & 911 $\pm$ 21  & 782 $\pm$ 27 \\

Agent-S3 w/ GPT-5~\cite{gonzalezpumariega2025unreasonableeffectivenessscalingagents}
& Agentic Framework
& 998 $\pm$ 21 & 1231 $\pm$ 44 & 623 $\pm$ 55  & 628 & 702 $\pm$ 34 & 633 $\pm$ 19 \\

\textbf{Human Expert}
& Human Agent
& \textbf{18 $\pm$ 5} & \textbf{25 $\pm$ 8} & \textbf{16 $\pm$ 3} & \textbf{48 $\pm$ 30} & \textbf{22 $\pm$ 9} & \textbf{11 $\pm$ 3} \\

\midrule
\textbf{OSExpert (Explore w/ Qwen-3-VL-8B, wo/ Fast Planner)}
& Agentic Framework
&  46 $\pm$ 1 &  31 $\pm$ 0  & 34 $\pm$ 2  & 49 $\pm$ 3 & 38 $\pm$ 0 & 44 $\pm$ 2 \\
\textbf{OSExpert (Explore w/ Qwen-3-VL-8B, wo/ Boundary Check)}
& Agentic Framework
&  84 $\pm$ 9 &  91 $\pm$ 11 & 75 $\pm$ 5 & 84 $\pm$ 11 & 61 $\pm$ 4  & 82 $\pm$ 3 \\
\textbf{OSExpert (Explore w/ Qwen-3-VL-8B)}
& Agentic Framework
&  \underline{32 $\pm$ 0} &  29 $\pm$ 2 & \underline{28 $\pm$ 0}  & \underline{41 $\pm$ 3} & \underline{35 $\pm$ 0}  & 39 $\pm$ 1 \\
\textbf{OSExpert (Exploration w/ GPT-5 \& UI-TARS)}
& Agentic Framework
& 34 $\pm$ 2 &  \underline{27 $\pm$ 3} & \underline{28 $\pm$ 0}  & 44 $\pm$ 2 & \underline{35 $\pm$ 0}  & \underline{37 $\pm$ 1} \\

\bottomrule
\end{tabular}}
\end{table*}

\subsection{Main Results}

\paragraph{Limited Performance and Efficiency with Current CUAs}

Table~\ref{tab:method_compare_performance} and Table~\ref{tab:method_compare_efficiency} reveal a large gap between current computer use agents and human experts on the complex tasks introduced by our OSExpert-Eval. Agent capability is strongly bounded by task complexity and by the environments and interfaces observed during training. On unseen UIs and fine-grained action execution, none of the agents exhibits meaningful generalization, with success rates remaining in the 0\% to 10\% range. In these settings, agents often misinterpret UI elements, become confused by novel screens, and either loop through incorrect actions or rely on trial and error without discovering the correct procedure. Although most agents perform well on OSWorld, where evaluation often emphasizes unit function tests and relatively simple tasks, their success rates drop sharply on long horizon workflows due to cascading errors. Trajectory analysis further suggests two primary sources of inefficiency. First, high per step latency: agents typically generate a new plan from the current observation at each step and then produce a single action, and this one step planning is repeated after every UI change. Second, reliance on test time scaling: agents persist through repeated failures on difficult steps and often terminate only when the interaction budget is exhausted. Overall, compared to human experts, current agents exhibit less stable performance and require about 5× to 50× longer elapsed time on OSExpert-Eval tasks. While the exact ratio depends on the iteration cap, stochasticity, and task difficulty, the latency gap remains substantial, and the performance gap persists even with additional interaction budget.

\paragraph{Performance and Efficiency Gain with OSExpert} Table~\ref{tab:method_compare_performance} and Table~\ref{tab:method_compare_efficiency} show that our environment learning framework yields substantial gains on OSExpert Eval. Starting from near zero performance for current computer use agents, OSExpert increases success rates to roughly 20\% to 30\% using only automated exploration, without additional human effort on curating exploration failures. Models equipped with the learned skill set also execute tasks more efficiently, reducing overall completion time and closing a large portion of the efficiency gap to human experts. These results highlight the potential of our environment learning framework for constructing next generation computer use agents.

%\paragraph{Efficiency Gain with OSExpert} According to Table~\ref{tab:method_compare_efficiency}, we observe that most agentic framework are significantly slower than end to end computer use models (either a general or specific model)

\subsection{Ablation Study and Qualitative Analysis} 

\paragraph{Gains from Environment Specific Knowledge} We find that the gains on OSExpert-Eval subsets targeting long horizon compositional workflows and unseen UI generalization primarily come from environment specific procedural knowledge stored in the skill set. On these tasks, existing computer use agents often rely on step wise planning without procedural knowledge, which makes them vulnerable to early mistakes that propagate across later steps. Such errors can trigger misguided exploration and poor recovery behavior, and agents may fail to return to a valid earlier state even after detecting the mistake. In contrast, our agent equipped with procedure skills of the specific digital environment executes verified plans and action templates from the skill set, so each step is more likely to be correct and the probability of entering unrecoverable states is reduced. These results underscore the importance of environment specific skills for robust multi-step execution. While we retrieve these skills using text embeddings, they could also be integrated into the reinforcement learning stage so that the agent can fully internalize them. We leave this exploration to future work.

\paragraph{Gains from Skills for Fine-Grained Actions} As shown in Table~\ref{tab:method_compare_performance}, We find that the gains on OSExpert-Eval subsets targeting fine-grained action execution mainly come from the skills constructed in Section~\ref{sec:3.3}. While current computer use agents are not trained to reliably perform these precise manipulations, our exploration procedure allows the agent to invoke action primitives from the database and succeed during environment learning. The resulting condensed skills then guide the agent to reproduce these behaviors at inference time. These results suggest an alternative to continuously scaling manual annotation or relying on complex end to end training for fine-grained control tasks such as tracing object boundaries, and point to a practical path toward human expert level manipulation.

\paragraph{Efficiency Gains from Fast Planner \& Skill Boundary Check} Table~\ref{tab:method_compare_efficiency} shows that our fast planner improves the overall efficiency of even the most efficient computer-use agent. From the table, we observe that most of the efficiency gain comes from the skill boundary check, with a smaller contribution from the fast planner. This is mainly because current computer-use agents rely heavily on test-time scaling and often spend excessive time attempting tasks that are unlikely to succeed until the maximum interaction limit is reached (30 steps in our setting). In real-world scenarios, the efficiency gain from knowledge boundary awareness would likely become more significant when longer interaction horizons are allowed. This suggests that deploying computer-use agents with stronger knowledge awareness can substantially improve practical efficiency. Although the fast planner currently provides a relatively small gain of about $\sim$10 seconds, its impact may become more pronounced in real-world tasks that involve longer horizons.

\section{Relatd Work}

\paragraph{General Computer Use Agents} With recent advances in LLMs and MLLMs, agents for digital environments have progressed from text-only interactive worlds~\cite{cote2018textworld,hausknecht2019jericho} and simulated benchmarks~\cite{liu2018wge,yao2022webshop} to operating in realistic web interfaces~\cite{zhou2023webarena,koh-etal-2024-visualwebarena} and real-world desktop and mobile applications~\cite{bonatti2024windowsagentarena,rawles2024androidworld,xie2024osworld,davydova2025osuniverse}. Trained on human demonstrations, either carefully annotated datasets or in-the-wild online video sources, these agents learn to perceive and interpret the screen, generate natural-language plans and reasoning, and ground them into executable action sequences~\cite{wang2025opencua,song2025coact1,wang2025uitars2,agashe2025agents2}. While their performance on daily, simple tasks is approaching human levels~\cite{gonzalezpumariega2025unreasonableeffectivenessscalingagents}, their efficiency often still lags behind and remains insufficient for practical real-world deployment~\cite{abhyankar2025osworldhuman}, and their ability to generalize to complex professional workflows remains limited. OSWorld-Human~\cite{abhyankar2025osworldhuman} evaluates the efficiency of computer-use agents on OSWorld~\cite{xie2024osworld}, showing that agents can incur substantial latency, particularly in the planning stage. OSUniverse~\cite{davydova2025osuniverse} stratifies tasks by complexity and reports that current agents struggle to solve the hardest tier. Recently, Anthropic advocated “Do not build agents, build skills instead”~\cite{zhang2025dontbuildagents}, suggesting that scaling training data for base models may not be the optimal way to construct general agents. In this work, we introduce automatically constructing a skill set by deploying computer use agents in the target digital environment.

\paragraph{Self-Evolving GUI Agents}
The vast and constantly changing nature of GUI environments makes exhaustive human annotation infeasible, self-evolving GUI agents~\cite{fang2025webevolver,zhang2025uievolautomaticknowledgeevolving,li2025appagentv2advancedagent,wang2025mobileagenteselfevolvingmobileassistant,sun2025seagent} therefore aim to autonomously acquire application-specific knowledge from interaction~\cite{gao2025survey}. WebEvolver coevolves a world model for better self-training and look-ahead planning~\cite{fang2025webevolver}, UI-Evol retraces interaction traces and critiques/refines externalized knowledge to better align with execution~\cite{zhang2025uievolautomaticknowledgeevolving}, and Mobile-Agent-E distills experience into reusable Tips and executable Shortcuts for complex mobile tasks~\cite{wang2025mobileagenteselfevolvingmobileassistant}. They leverage exploration differently: Most works like AppAgent-v2 build UI knowledge and uses retrieval-augmented execution~\cite{li2025appagentv2advancedagent}, while SEAgent continually updates skills from experience via reinforcement learning~\cite{sun2025seagent}. In contrast, our environment-learned agents learn comprehensively from unit functions rather than user-query-driven exploration, and introduce procedural skill caching for efficient planning and fine-grained tool creation toward expert-level performance.

%The vast and constantly changing nature of GUI environments makes exhaustive human annotation infeasible, and self-evolving gui agents~\cite{fang2025webevolver,zhang2025uievolautomaticknowledgeevolving,li2025appagentv2advancedagent,wang2025mobileagenteselfevolvingmobileassistant,sun2025seagent} have emerged as a promising approach to autonomously acquire application-specific knowledge~\cite{gao2025survey}. WebEvolver coevolving a world model and supports look-ahead for decision making \cite{fang2025webevolver}, and UI-Evol features critiquing and refining the externalized knowledge\cite{zhang2025uievolautomaticknowledgeevolving}. Mobile-Agent-E \cite{wang2025mobileagenteselfevolvingmobileassistant}.sumaaeizes useful tips and executable shortcuts for complex mobile tasks. These appraoches also utilize the exploration trajectories differently, with AppAgent-v2~\cite{li2025appagentv2advancedagent} primarily use retrieve augmented generation, SEAgent ~\cite{sun2025seagent} let the agent learn from this expeience with reinforcement learning. 

%~\cite{sun2025seagent}
%~\cite{li2025appagentv2advancedagent}

\section{Conclusion}

In this work, we propose a paradigm for developing more capable and general computer use agents by directly deploying them in digital environments and enabling them to autonomously construct skill sets through bottom up exploration. During this process, agents learn skills for executing fine-grained actions that require precise control, and their efficiency is further improved through a fast planner derived from the learned skills and through explicit awareness of their skill boundaries. Together, our algorithm and design choices enable agents to acquire professional level competencies, bringing general computer use agents a significant step closer to human expert level performance. 

By enabling agents to self-discover, compose, and reuse skills across environments, our approach also addresses a central bottleneck in building more general intelligent systems. Current agents are often tied to specific tasks, interfaces, or training distributions, which limits their applicability in open world scenarios. Paradigms like OSExpert, which emphasize environment driven skill discovery and compositional procedural knowledge, offer a promising way forward. While we focus on computer use agents, these principles extend naturally to software engineering agents, embodied robots, and vision language action models deployed on robots to assist humans. We hope OSExpert serves not only as a practical system but also as a step toward adaptive, next generation self improving agents and, ultimately, more general purpose intelligence.

%\heng{add discussion about limitations oand future work}
%\jiateng{added below and future work in appendix}

\section*{Impact Statement}
This paper aims to advance machine learning by improving the capability, efficiency, and robustness of computer use agents through environment driven skill discovery. If deployed responsibly, such agents may increase productivity in professional software workflows by assisting users with complex, multi step tasks. As with other automation systems, risks include overreliance, unexpected failures in novel interfaces, and misuse in settings where errors are costly. Our approach mitigates some of these risks by constructing a verifiable skill set through bottom up exploration, organizing successful procedures into reusable skills and recording persistent failures as explicit boundaries. We also use a fast planner derived from the learned skills to reduce step wise planning overhead while retaining per step perception and action execution. For fine-grained GUI operations that require precise visual grounding, we leverage a database of action primitives and add a procedure to the skill set only when it succeeds under verification, reducing reliance on uncontrolled trial and error. Overall, we do not identify ethical concerns beyond those commonly associated with deploying machine learning based automation and decision support systems, but we emphasize the importance of human oversight and cautious use in high stakes environments.

\section*{Limitations}

While OSExpert demonstrates substantial improvements in both performance and efficiency over existing computer use agents, several limitations remain.

\textbf{Exploration cost and scalability.}
Our environment learning framework relies on explicit interaction with the target digital environment, including repeated environment resets, action replays, and verification. Although GUI-DFS is designed to yield verified skills early at training stage, exploration can still be time and compute intensive for large applications with deeply nested menus, high branching factors, or highly combinatorial interfaces. Scaling OSExpert to environments with hundreds of tools, dynamic UI layouts, or frequent version changes remains an open challenge. We discuss about our current exploration cost with different models and conditions in Appendix~\ref{app:Details}.

\textbf{Dependence on base model capabilities.}
The quality of the learned skill set depends on the underlying planner, action module, and feedback module used during exploration. Stronger models may discover more reliable skills and recover from difficult states more effectively, while weaker models may produce incomplete or noisy skill sets. Although OSExpert avoids environment specific human demonstrations, it does not eliminate reliance on strong foundation models, and certain amount of human annotation to provide verifiable hint of the exploration stage and revise the skills later could still be extremely vital and helpful in real-world deployments.

\textbf{Incomplete coverage and brittle failure modes.}
Under finite budgets, GUI-DFS does not guarantee exhaustive coverage of all valid interaction paths. Some rare or deeply nested functionalities may remain undiscovered, and certain interactions may still fail due to visual ambiguity, occlusions, non deterministic UI behavior, or unexpected pop ups. While failed exploration traces are recorded and used for early stopping, they may reflect limitations of the exploration policy rather than true infeasibility.

\textbf{Manual design of fine-grained primitives.}
Our fine-grained action handling requires identifying recurring manipulation patterns and providing corresponding action primitives and grounding tools. Although this effort is substantially smaller than large scale annotation or end to end supervised training, it still introduces manual design and domain knowledge. Automating the discovery, verification, and refinement of such primitives is an important direction for future work.

\textbf{Evaluation scope.}
OSExpert-Eval focuses on professional desktop applications and GUI-based workflows. Our evaluation is conducted on the OSExpert-Eval benchmark and OSWorld (in Appendix~\ref{app:data detail}), an open-source benchmark covering six different digital environments. While these benchmarks capture challenges that are underrepresented in existing evaluations, they do not cover all interaction settings, such as mobile interfaces, purely web-native automation, or continuous real-time environments. Extending environment learning and skill construction to these diverse application settings is left for future work.

Overall, these limitations reflect trade offs inherent in environment driven learning. We view OSExpert as a step toward more adaptive and self improving computer use agents, and we believe addressing these challenges will further advance the field.

%\section*{Limitations}

%\subsection*{Boundaries of Environment Learning}


\bibliography{example_paper}
\bibliographystyle{icml2026}

%%%%%%%%%%%%%%%%%%%%%%%%%%%%%%%%%%%%%%%%%%%%%%%%%%%%%%%%%%%%%%%%%%%%%%%%%%%%%%%
%%%%%%%%%%%%%%%%%%%%%%%%%%%%%%%%%%%%%%%%%%%%%%%%%%%%%%%%%%%%%%%%%%%%%%%%%%%%%%%
% APPENDIX
%%%%%%%%%%%%%%%%%%%%%%%%%%%%%%%%%%%%%%%%%%%%%%%%%%%%%%%%%%%%%%%%%%%%%%%%%%%%%%%
%%%%%%%%%%%%%%%%%%%%%%%%%%%%%%%%%%%%%%%%%%%%%%%%%%%%%%%%%%%%%%%%%%%%%%%%%%%%%%%

%%%%%%%%%%%%%%%%%%%%%%%%%%%%%%%%%%%%%%%%%%%%%%%%%%%%%%%%%%%%%%%%%%%%%%%%%%%%%%%
%%%%%%%%%%%%%%%%%%%%%%%%%%%%%%%%%%%%%%%%%%%%%%%%%%%%%%%%%%%%%%%%%%%%%%%%%%%%%%%

\newpage
%\onecolumn
\appendix

\section{Additional Experimental Results}
\label{app:Details}

%\subsection{Ablation Studies}
%In this section, we conduct a series of experiments to evaluate the standalone contribution of each component in our environment-learning framework for building expert computer-use agents. Specifically, we isolate the effect of applying GUI-DFS (Section~\ref{sec:3.1}) without the fast planner (Section~\ref{sec:3.2}) or the fine-grained skill construction module (Section~\ref{sec:3.3}). We further quantify the efficiency gains attributable to the planner and to the skill boundary check, respectively.

\subsection{Performance on General CUA Benchmarks}
\label{app:general}

We further evaluate the effectiveness of our agent equipped with GUI-DFS on a general computer-use agent benchmark, particularly OSWorld~\cite{xie2024osworld}. As shown in Table~\ref{tab:osworld_subsplits}, our method consistently outperforms the baseline across the evaluated subsets, although the margin of improvement is relatively modest. This behavior is expected because GUI-DFS primarily mitigates error propagation by enabling more structured exploration and recovery during task execution, rather than directly improving single-step knowledge or perception capabilities. Given the limited number of retries allowed during the exploration stage, the models are hardly able to go beyond their knowledge boundary and explore the knowledge that they could not cover given the current priors. As a result, the gains mainly arise from improved robustness in multi-step decision making rather than from enhanced step-level competence. 

\begin{table}[t]
\centering
\small
\begin{tabular}{lcccc}
\toprule
\textbf{Model} & \textbf{GIMP} & \textbf{Calc} & \textbf{Impress} & \textbf{Writter} \\
\midrule
Base Agent & 0.269 & 0.447 & 0.404 & 0.304 \\
\textbf{OSExpert} & 0.308 & 0.447  & 0.426 & 0.347 \\
\bottomrule
\end{tabular}
\caption{Performance Comparison between the Base Computer Use Agent with Qwen-3-VL-8B and the our agent equipped with Exploration results on four OSWorld subsets using Pass@1 Success Rate ($\uparrow$).}
\label{tab:osworld_subsplits}
\end{table}

\subsection{Computational Budget Analysis}

\paragraph{Computational Budget Analysis.}
The computational cost of the exploration process in \textbf{GUI-BFS} is primarily determined by the number of nodes explored in the interface state space. Suppose that from the initial state there exist $N$ distinct functions (or targets) that the agent may attempt to reach, and that each function requires on average $M$ interaction steps to reach. In this case, the exploration cost grows proportionally to the total number of visited nodes, leading to a base complexity of $O(M \times N)$.

In practice, the model may occasionally hallucinate additional branches during exploration. Let $K$ denote the hallucination rate, representing the proportion of additional branches expanded due to incorrect or redundant actions. These expansions introduce an additional linear factor in the number of explored nodes. However, since we explicitly cap the maximum exploration depth for each candidate function, the hallucinated branches cannot grow unbounded and therefore only incur a bounded linear overhead.

Consequently, the overall computational complexity remains linear in the number of explored nodes. If we denote the average computation time per node as $T_{\text{node}}$, which includes the planner, action execution, and environment feedback stages, the total computational budget of the exploration process can be approximated as
\[
O(M \times N \times T_{\text{node}})
\]
This analysis indicates that the exploration cost scales linearly with the number of candidate functions and the interaction depth required to reach them, while hallucinated branches introduce only a bounded constant-factor overhead due to the depth constraint.

As a concrete example, the full exploration of our \textsc{MiniWord} simulated environment involves visiting 415 nodes. For larger-scale environments such as GIMP and the LibreOffice suite, we restrict the exploration to subsections that are most relevant to downstream tasks. For instance, in GIMP we focus on functions under \textit{Image}, \textit{Colors}, and several commonly used tools, while in LibreOffice we concentrate on the interactive panel and a subset of frequently used editing tools. By constraining the initial exploration space in this manner, we significantly reduce the number of functions requiring exploration, keeping the total number of explored nodes at the scale of hundreds rather than thousands. In practice, this results in 721 explored states for GIMP and 210 states for LibreOffice Writer. Overall, it costs around 20\$ to 50\$ for exploration on each digital environment for our experiments with GPT-5 models. However, conducting a complete exploration of real-world environments may incur substantially higher computational cost. We therefore recommend leveraging stronger foundation models for future large-scale and comprehensive exploration.

\section{OSExpert-Eval Benchmark Details}
\label{app:data detail}

\begin{figure}[t]
    \centering
    \includegraphics[width=1.0\columnwidth]{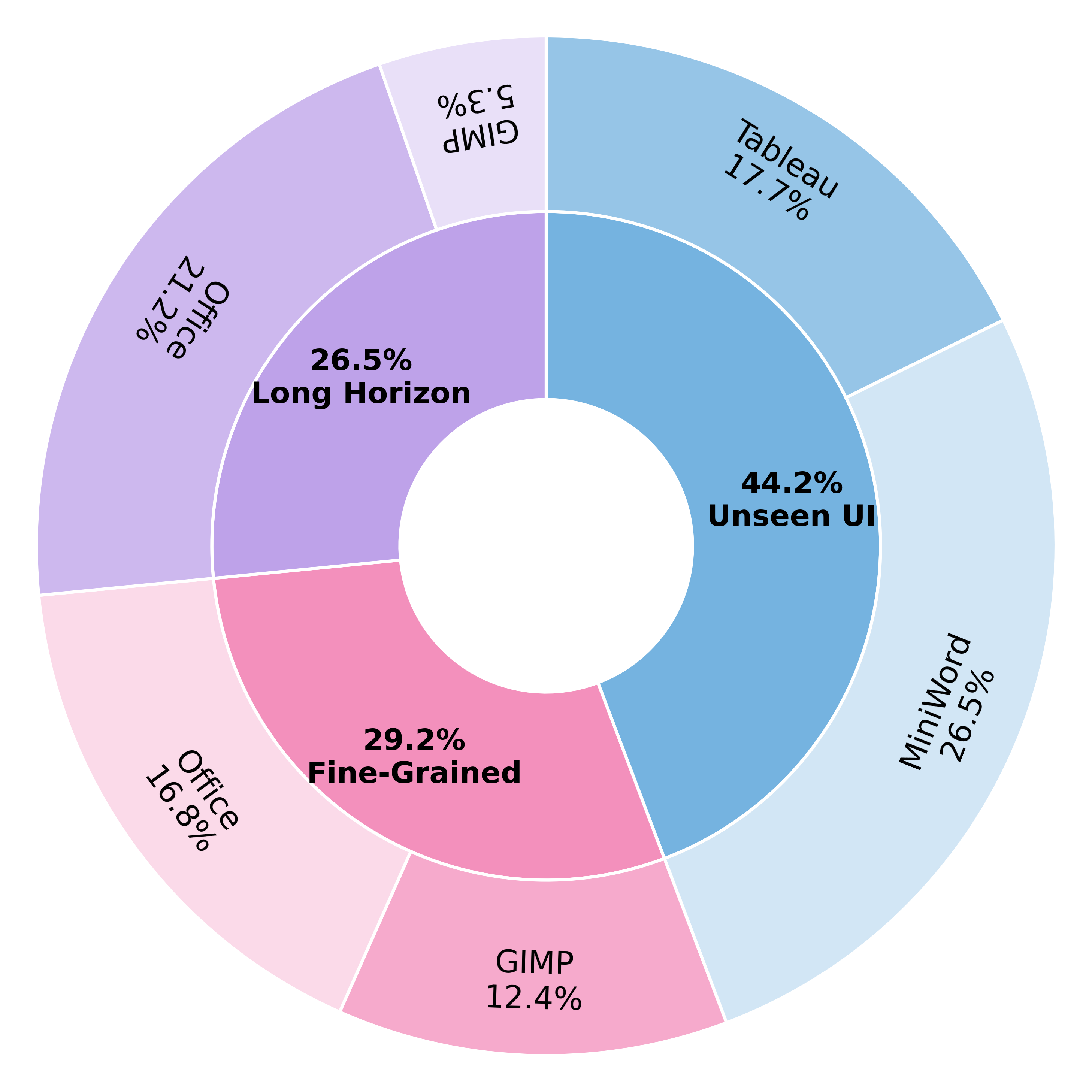}
    \caption{Composition of our evaluation tasks (113 total). The inner ring shows the three high-level categories (Unseen UI, Fine-Grained, and Long Horizon), while the outer ring breaks each category down by environment (Tableau, MiniWord, Office, and GIMP); slice sizes are proportional to the number of tasks. Here, Office includes LibreOffice Writer, LibreOffice Impress, and LibreOffice Calc.}
    \label{fig:data}
    \vspace{-6pt}
\end{figure}

OSExpert-Eval is a curated benchmark designed to evaluate computer-use agents on professional-level challenges that go beyond routine GUI interaction. The benchmark consists of 113 tasks spanning three major categories: \emph{Long-Horizon Compositional Workflows}, \emph{Unseen UI Generalization}, and \emph{Fine-Grained Action Execution}. The overall task composition and environment breakdown are summarized in Figure~\ref{fig:data}.  

Specifically, \textbf{Long Horizon contains 30 tasks}, including 24 Office tasks and 6 GIMP tasks, emphasizing multi-step workflows that require composing multiple unit functions in a correct and robust order. \textbf{Unseen UI contains 50 tasks}, including 20 Tableau tasks and 30 MiniWord tasks, targeting novel layouts and interaction patterns that are uncommon in current agents' training distributions. \textbf{Fine-Grained contains 33 tasks}, including 14 GIMP tasks and 19 Office tasks, requiring precise low-level control such as accurate text selection, object manipulation, and spatial alignment.

\begin{figure*}[t]
    \centering
    \includegraphics[width=2.0\columnwidth]{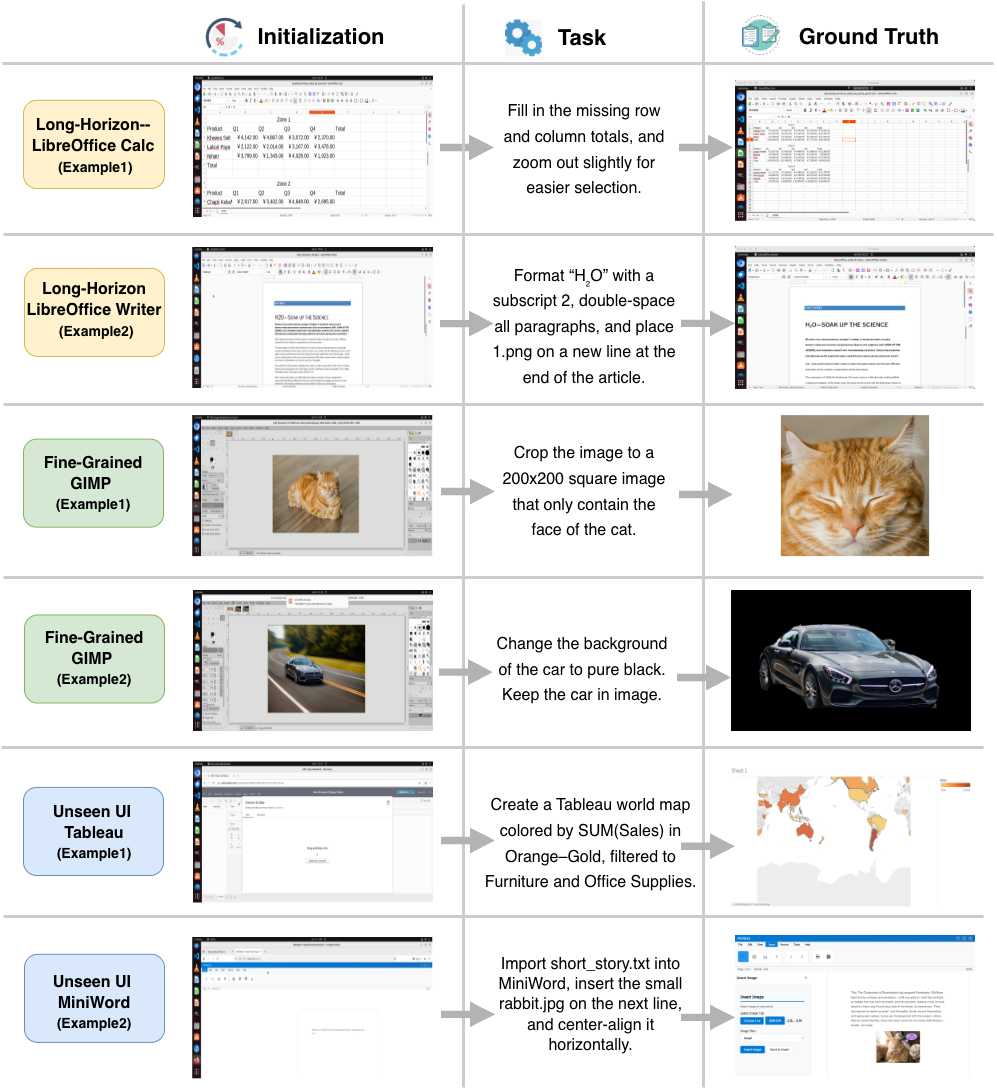}
    \caption{Representative examples from OSExpert-Eval across three task categories. The figure aggregates six examples illustrating the breadth of professional computer-use skills evaluated in our benchmark. \textbf{Top (Long-Horizon Compositional Workflows)}: multi-step tasks in LibreOffice Calc and Writer that require composing several unit operations in the correct order, including spreadsheet completion with interface adjustments (e.g., zoom) and document-wide formatting with image insertion. \textbf{Middle (Fine-Grained Action Execution)}: precise image-editing tasks in GIMP, including tightly cropping to a specified 200×200 region and performing accurate background removal while preserving object integrity. For each example, we show the initial environment state, the natural-language instruction, and the corresponding ground-truth outcome. \textbf{Bottom (Unseen UI Generalization)}: tasks in Tableau and MiniWord that test transfer to unfamiliar interfaces and interaction patterns, such as building a world map visualization with sales-based coloring and category filtering, and importing external text followed by image insertion and alignment in a novel editor layout.
}
    \label{fig:examples}
    \vspace{-6pt}
\end{figure*}

Below, we present representative examples from each category, including the task instructions and their corresponding ground-truth outcomes (Figures~\ref{fig:examples}).

The dataset was manually constructed by the paper’s coauthors, who each spent over two hours becoming familiar with the core functionalities of the target software. After designing the tasks, they executed and recorded their own task completion trajectories to establish reference performance. We will release the full dataset and evaluation code upon acceptance. 

\section{Algorithm Design: GUI-BFS or GUI-DFS}
\label{sec:appendix_bfs_dfs}

%\begin{comment}
We discuss the trade offs between the GUI-DFS procedure used in Algorithm~\ref{alg:gui_DFS_reverse} and a GUI-BFS (Algorithm~\ref{alg:gui_BFS_reverse}) variant that follows the same exploration interface but replaces the stack with a queue. Both strategies are sound in the sense that, given sufficient budget and the same planner, action module, and feedback module, they can eventually enumerate comparable exploration targets and yield a unit function skill set. The practical differences arise from the constraints of computer use scenarios, where each node expansion requires an environment restart, action replay, and verification, and where interaction budgets, latency, and memory are limited. Note that We ensure that both algorithms are reaching to a safe end by limiting the maximum depth of any single exploration node.

\paragraph{Time to first useful skills.}
A key advantage of GUI-DFS is fast skill yield. DFS commits to one branch of the interface hierarchy and quickly reaches terminal states where a unit function can be verified and condensed into a reusable skill. This matters in computer use settings because skills are valuable as soon as they are discovered. Even if exploration is terminated early due to time limits or a stopping rule, DFS typically returns a nontrivial set of completed procedures and action templates that can already improve downstream performance. In contrast, GUI-BFS prioritizes breadth and spends early budget expanding many shallow nodes. This can delay reaching terminal states, resulting in fewer finalized skills (or leaf nodes explored) under the same exploration budget.

\paragraph{Memory and frontier growth.}
GUI-BFS maintains a wide frontier of intermediate nodes, which can grow rapidly with the branching factor of GUI menus, toolbars, and dialogs. In practice, this implies higher memory usage for storing exploration nodes and their associated plans and action prefixes. GUI-DFS keeps only a single active path plus a smaller set of deferred siblings, making its peak memory substantially lower in typical GUI trees. This difference is especially pronounced in professional software where top level menus expose many options and each option can further branch into dialogs and subpanels.

\paragraph{Summary.}
For computer use agents, we generally prefer GUI-DFS, as used in Algorithm~\ref{alg:gui_DFS_reverse}, because it produces verified unit skills quickly, supports useful partial results under early stopping, and has lower peak memory due to a smaller frontier. GUI-BFS is a reasonable alternative only when the primary objective is broad coverage of shallow UI functions and sufficient budget is available.
%\end{comment}

\begin{algorithm}[t]
\caption{GUI-BFS Algorithm
}
\label{alg:gui_BFS_reverse}

{\small
\textbf{Definitions.}
Digital environment $E$; Environment State $S_i$; Exploration Plans $\Pi$; Action Sequence $\alpha$;
Exploration State Node
$n \triangleq (\Pi,\alpha)$; Exploration Outcome $T \in \{\mathtt{Continue},\mathtt{Final},\mathtt{Error}\}$.

\par\textbf{Inputs.}
Planner Module $P$; Action Module $A$; Feedback Module $F$; Max Retries $R$.

\textbf{Output.}
Unit Skill Set $\mathcal{K}$ for Environment $E$.
\par}

\begin{algorithmic}[1]
\STATE $\mathcal{Q} \leftarrow \emptyset,\ \mathcal{K} \leftarrow \emptyset$, $(S_0) \leftarrow \texttt{Reset}(E)$
\STATE $\mathcal{Q} \leftarrow \texttt{EnqueueAll}(\mathcal{Q},\ P(E,S_0))$ 
\WHILE{$\mathcal{Q} \neq \emptyset$}
  \STATE $(\Pi,\alpha) \leftarrow \texttt{Dequeue}(\mathcal{Q})$; $(S_0) \leftarrow \texttt{Reset}(E)$
  \STATE $S_i \leftarrow \texttt{Execute}(E,\alpha)$; $r \leftarrow 0$
  \WHILE{$r < R$}
    \STATE $\alpha' \leftarrow A(\Pi,\alpha,S_i)$; $S_{i+1} \leftarrow \texttt{Execute}(E,S_i,\alpha')$
    \STATE $(T,feedback)\leftarrow F(S_i,S_{i+1},\Pi,\alpha')$
    \IF{$T=\mathtt{Continue}$}
      \STATE $\mathcal{K} \leftarrow \mathcal{K} \cup \{(\Pi,\ \alpha \oplus \alpha')\}$
      \STATE \textbf{for} each $\Pi^{new} \in P(E,S_{i+1},\Pi,\alpha \oplus \alpha')$ \textbf{do}
      \STATE \hspace{1em}$\mathcal{Q} \leftarrow \texttt{Enqueue}\!\left(\mathcal{Q},\ (\Pi \oplus \Pi^{new},\ (\alpha \oplus \alpha') )\right)$
      \STATE \textbf{end for}
      \STATE \textbf{break}
    \ELSIF{$T=\mathtt{Final}$}
      \STATE $\mathcal{K} \leftarrow \mathcal{K} \cup \{(\Pi,\ \alpha \oplus \alpha')\}$
      \STATE \textbf{break}
    \ELSE 
      \STATE $\mathcal{Q} \leftarrow \texttt{Enqueue}\!\left(\mathcal{Q},\ (\Pi \oplus feedback,\ \alpha \oplus \alpha')\right)$
      \STATE $r \leftarrow r + 1$
      \STATE \textbf{break}
    \ENDIF
  \ENDWHILE
  \IF{$r \ge R$}
    \STATE $\mathcal{K} \leftarrow \mathcal{K} \cup \{(\Pi,\ \alpha)\}$
  \ENDIF
\ENDWHILE
\STATE \textbf{return} $\mathcal{K}$
\end{algorithmic}
\end{algorithm}

%\newpage

\section{Discussion and Future Work}

In this section, we discuss several insights from our work and highlight opportunities that it points to as promising directions for future research.

\paragraph{Difference between Exploration and Test-Time Scaling.}
Exploration can be viewed as a form of scaling that shifts computation from inference time to a one-time environment learning phase. While test-time scaling repeatedly explores alternative trajectories during inference, self-exploration amortizes this cost by converting accumulated experience into reusable procedural knowledge before the system is deployed and interacts with users. Both forms of scaling share several common challenges while also exhibiting distinct trade-offs that influence overall response quality. The shared challenges include diminishing returns with increasing attempts, long-tail failure cases, and the difficulty of determining when additional attempts are unlikely to produce better outcomes. In terms of trade-offs, our skill boundary check provides an initial step toward principled early stopping, which benefits users by preventing excessive waiting time caused by repeated inference attempts. At the same time, the exploration stage enables the construction of a more comprehensive set of skills, making the agent generally more capable once deployment begins. However, exploration itself typically requires more time than the query-and-serve mode (where a user query arrives and the system begins test-time scaling immediately), and it may also introduce additional exploration errors due to the absence of a clearly specified target during exploration. Overall, our study highlights the importance of understanding how to allocate exploration budgets and how to balance exploration with test-time scaling. Determining when each paradigm should be preferred remains an important direction for future research.

\paragraph{Resolving Structuring Complexity and Nested Interfaces.}
Modern GUIs often expose deeply nested functionalities through menus, dialogs, toolbars, and mode-dependent panels. Identifying functional boundaries and hierarchical structures within such interfaces remains challenging for current foundation models, whose performance still heavily relies on heuristics such as carefully designed initial exploration stages or prompt engineering. We believe that the generalization capability of such frameworks will improve significantly as stronger base models become available in the future.

\paragraph{Compositionality and Skill Abstraction.}
Many professional workflows arise from nontrivial compositions of atomic UI operations. While our curriculum-based extension allows agents to construct composite skills from unit functions, the combinatorial space of possible compositions grows rapidly with environment complexity. Future work could explore more principled abstractions for skill composition, such as hierarchical skill graphs or program induction methods that reason over reusable subprocedures rather than enumerating combinations.

\paragraph{Skill Representation and Knowledge Transfer.}
In OSExpert, skills are stored as verified plans and action templates tied to specific environments. A natural extension is to study how such procedural knowledge can be represented more abstractly and transferred across related applications, interface versions, or task domains. Learning environment-agnostic representations of skills, while preserving their verifiability and grounding, would move computer-use agents closer to generalizable competence.

\paragraph{Human Agent Collaboration in Exploration.}
Although our framework emphasizes fully autonomous exploration, human input remains a valuable source of guidance, especially for rare failure cases or ambiguous user-interface behaviors. Future systems could support mixed-initiative exploration, where agents propose candidate skills and humans provide lightweight feedback, correction, or validation. Such collaboration could further improve skill quality while maintaining low annotation cost. A scalable way for organzing the learned skills for human and enable easy double check and revision of the skills should also be of significant importance at the current stage. 

\paragraph{Beyond GUI-Based Agents.}
Although this work focuses on desktop GUI environments, the underlying principles of environment-driven skill discovery, verifiable procedural knowledge, and boundary-aware execution extend naturally to broader settings. These include vision–language–action models operating in embodied environments, robotics systems interacting with physical tools, and software engineering agents navigating code editors and development workflows. Exploring these extensions may help unify environment learning across digital and physical domains.

Overall, we view OSExpert as a step toward shifting agent design from reactive test-time scaling toward proactive, structured, environment-grounded learning. While there are still challenges in the way, we believe that this framework may enable more adaptive, efficient, and trustworthy agents across a wide range of interactive settings.

\section{Prompts Used During Exploration}

This section documents the prompts used by different modules during environment exploration. We separate prompts by functional role, including file preparation, planning, and feedback. The planning prompts guide the agent in proposing exploration targets and interaction strategies, while the feedback prompts are responsible for classifying state transitions, verifying terminal outcomes, and diagnosing failures. All prompts are designed to be lightweight, modular, and reusable across environments, and are included to support transparency and reproducibility of the exploration process.

% ---------- File preparation planner ----------
\begin{FilePrepPromptBox}
% TODO: paste the file-preparation planner prompt here.
Your job is to decide whether this application needs a specific example file or document to be ready before deeper exploration can proceed. \\

Based ONLY on what you can see in the screenshot and the exploration goal, you must answer: \newline 

  \textbf{- 1)} Does the agent need to open or create a concrete file/document/data object to meaningfully explore this application? \newline
  
  \textbf{- 2)} If yes, what is the simplest, most representative file or object type that should be used? \newline
  
  \textbf{- 3)} Should the agent:
     - create a new simple example file (``self-create-file''),
     - load a human-provided example file (``use-provided-file''), or
     - proceed without preparing a file because there is already sufficient content visible (``no-file-needed'')? \newline

\textbf{Exploration Goal:} \textit{\{instruction\}} \newline

You MUST output a JSON object with the following fields: \newline

\{\{
  ``file-decision'': ``one of `self-create-file', `use-provided-file', or `no-file-needed'", \newline
  
  ``justification-decision": ``Short explanation of why this choice is appropriate given the visible UI", \newline
  
  ``suggested-file-type": ``If a file is needed, what is the simplest representative type (e.g., `short text document', `2-column CSV with dummy data', `small image file'); otherwise use `none'", \newline
  
  ``justification-alternatives": ``Very brief explanation of why the other options are less suitable in this specific UI state" \newline
  
\}\} \newline

\textbf{Guidelines}: \newline 

- If the UI clearly shows an empty document area with tools for editing, and no specific file content is required, it is often acceptable to choose ``no-file-needed". \newline

- If the UI suggests typical ``File/Open" or ``File/New" workflows and there is no existing content, strongly consider ``self-create-file" with a minimal example (e.g., one short paragraph, a tiny table). \newline

- Use ``use-provided-file" when the UI or goal explicitly suggests working with a specific sample dataset or external document (e.g., ``open the provided report", ``analyze the given CSV"). \newline

- Always keep your justifications short and grounded in what is visible. \newline

Provide your response in JSON format only.
\end{FilePrepPromptBox}

% ---------- Planning: initialize exploration targets ----------
\begin{InitExplorePromptBox}
You are an INITIAL EXPLORATION PLANNER analyzing a completely new and unknown software application called \textit{\{app-name\}.} \newline

  \textbf{CRITICAL INSTRUCTIONS:} \newline
  
  1. You may use your general knowledge about software interfaces, but your analysis must be grounded in what is visible in the screenshot.\newline
  
  2. Do not hallucinate UI elements or behaviors that are not supported by visible evidence in the screenshot.\newline
  
  3. Treat this as if you are seeing this specific software interface for the very first time. \newline
  
  4. Focus ONLY on the {app-name} application window. Ignore operating system elements like the top OS bar, clock, calendar, system notifications, or window controls (minimize/maximize/close buttons). Only analyze the application's own UI elements.\newline

  \textbf{Exploration Goal:} \textit{  \{instruction\}} \newline

  Current Step (Stage 1 - initial exploration): \textit{\{step-idx\}}

  Recent Initial-Exploration History:
  \textit{\{history-text if history-text else `No initial exploration history yet'\}} \newline

  Analyze the \textit{\{app-name\}} application window by carefully examining the screenshot. \newline

  \textbf{MAIN OBJECTIVE:} \newline
  (Stage 1 - Initial Screen Understanding): \newline
  
  - Infer, based on the screenshot, what this software is probably used for. \newline
  
  - Infer what kinds of files or objects it most likely operates on (e.g., text documents, tabular data, images, project files, etc.). \newline
  
  - Propose a SMALL set of short exploratory actions that an action agent could perform to quickly test these hypotheses, with a focus on: * what kinds of data or files the application processes, and * how the main menus / side bars might be organized hierarchically. \newline

  You MUST output a JSON object with the following fields:
  \newline
  \{\{
  \newline
    ``application-hypothesis'': \{\{
    \newline
    \newline
      "likely-purpose": "Short description of what this software is probably used for, grounded in the screenshot",
      \newline
      
      ``possible-file-types'': [
        ``List of likely file or object types this software can open/edit/create (e.g., `text documents', `spreadsheets', `presentations')''
      ]
      \newline
      \newline
    \}\},
    \newline
    
    ``initial-probing-actions'': [
    \newline
    
      \{\{
      \newline
      
        ``description'': ``Short natural language description of what to try to better understand the software (e.g., `open the Edit menu and inspect available options')'',
        \newline
        
        ``instruction'': ``Concrete instruction for the action agent to execute in order to perform this probing action'',
        \newline
        
        ``target-area'': ``Optional: name or description of the UI area where this action should be applied (e.g., `top menu bar', `left toolbar')'',
        \newline
        
        ``expected-signal'': ``What this probing action should reveal about EITHER (a) what data/file types the application works with OR (b) how the main menus / side bars are organized hierarchically''

      \}\}

    ],
    \newline
    
    ``is-terminal'': false

  \}\}
  \newline

  \textbf{Guidelines:} \newline
  
  - The probing actions should be SHORT and focused, suitable for a single or small sequence of clicks/keystrokes. \newline
  
  - Prefer actions that reveal high-level information such as available file operations, document types, or major modes of the application. \newline
  
  - Use the recent initial-exploration history to avoid repeating actions that have already been tried. \newline
  
  - You may set ``is-terminal'' to true when you believe the initial exploration goal has been sufficiently achieved for this stage. \newline

  Provide your response in JSON format only.
\end{InitExplorePromptBox}

% ---------- Planning: initialize shallow UI interaction ----------
\begin{InitInteractPromptBox}
You are a high-level exploration planner analyzing a completely new and unknown software application called \textit{\{app-name\}}. \newline

  \textbf{CRITICAL INSTRUCTIONS:}\newline
  
  1. This is a BRAND NEW software that you have NO prior knowledge about. Do NOT use any knowledge about existing software interfaces (like GIMP, Photoshop, or any other applications). \newline
  
  2. You must base your analysis EXCLUSIVELY on what you observe in the screenshot. Do not make assumptions based on prior knowledge. \newline
  
  3. Treat this as if you are seeing this software interface for the very first time. \newline
  
  4. Focus ONLY on the \textit{\{app-name\}} application window. Ignore operating system elements like the top OS bar, clock, calendar, system notifications, or window controls (minimize/maximize/close buttons). Only analyze the application's own UI elements. \newline

  \textbf{Exploration Goal:} \textit{\{instruction\}} \newline

  \textbf{Current Step:} \textit{\{step-idx\}} \newline

  \textbf{Recent Exploration History:} 
  \textit{\{history-text if history-text else `No history yet'\}}
 \newline
 
  \textbf{Previous Stage 1 Summary}(if available): 
  \newline
  
  \textbf{- Application hypothesis: }
  \newline
  \textit{json.dumps(application-hypothesis) if  application-hypothesis is not None else 'N/A'}
  \newline
  
  \textbf{- Menu hierarchy summary:}

  \textit{menu-hierarchy-summary if menu-hierarchy-summary else 'N/A'}
  \newline

  \textbf{Analyze the \textit{\{app-name\}} application window by carefully examining the screenshot. } \newline

  \textbf{STEP 1:} Identify major UI areas by observing the screenshot layout. Name each area based on what you actually see - their location (top, left, right, bottom, center) and visual appearance (bar, panel, canvas, dock, toolbox, etc.). The area names should describe their location and content based purely on visual observation from the screenshot. 
  \newline

  \textbf{STEP 2:} Using BOTH the screenshot and the Stage 1 menu hierarchy summary (if provided), identify which interactive elements are:
  \newline
  
    - highest-level menus/controls that should each be the ROOT of a dedicated exploration (to go into ``exploration-plan''), and
    \newline
    
    - lower-hierarchy elements (submenus, toolbar buttons, nested controls) that will naturally be covered when exploring those root items (to go into ``lower-hierarchy-elements'').
    \newline
    
    When the Stage 1 hierarchy suggests that a region is context-dependent or subordinate (for example, a toolbar row whose contents change based on the active menu or mode), you MUST represent that region as lower-hierarchy under its controlling root(s), and NOT as its own root item in ``exploration-plan''.
    \newline

  \textbf{STEP 3}: For areas or elements that appear static, decorative, or unlikely to support meaningful interaction (e.g., logos, static backgrounds, non-clickable labels), list them separately in ``low-priority-areas'' with a brief explanation.
  \newline

  \textbf{CRITICAL CONNECTION:} 
  \newline
  
  The keys in "exploration-plan" MUST exactly match the ``area-name'' values from ``screen-areas-summary''. Each area you identify in screen-areas-summary should have a corresponding key in exploration-plan with its list of interactive elements. This ensures the hierarchical organization is consistent.
  \newline

  \textbf{Output format (JSON only):}

  \{\{
  \newline
  
    ``screen-areas-summary'': [
    \newline
    
      \{\{
      \newline
      
        ``area-name'': ``Name this area based on what you observe in the screenshot (location + visual appearance, e.g., `Top horizontal bar with text menus', `Left vertical panel with icon buttons', `Center canvas area showing image', `Right side panel with tabs')'',
        \newline
        
        ``description'': ``Brief natural language description of this area and its main purpose based on visual observation''
        \newline

      \}\}

    ],
    \newline
    
    ``exploration-plan'':
    \newline \newline
    \{\{

      ``[Area Name 1 - must exactly match area-name from screen-areas-summary]'': [
      \newline
      
        \{\{
        \newline \newline
          ``exploration-space'': ``Identifier for a HIGHEST-LEVEL menu/control or major functional region that should be explored as a root (e.g., a top menu like `File', a primary sidebar section, a major mode switch button)'',
          \newline \newline
          ``instruction'': ``ONE short, concrete next-step instruction for what the agent should do RIGHT NOW with this element (e.g., `Click the File menu', `Click the Insert tab'). Do NOT describe long multi-step procedures here.''

        \}\}
        \newline \newline
      ],
      \newline \newline
      ``[Area Name 2 - must exactly match area-name from screen-areas-summary]'': [
      \newline \newline
        \{\{
        \newline \newline
          ``exploration-space'': ``Identifier for another highest-level menu/control or functional region'',
          \newline \newline
          ``instruction'': ``ONE short, concrete next-step instruction for what the agent should do with this element based on the current observation."
          \newline \newline
        \}\}
        \newline \newline
      ]
      \newline \newline
    \}\},
    \newline \newline
    ``lower-hierarchy-elements'': [
      \{\{
      \newline
      
        ``area-name'': "Area this lower-hierarchy group belongs to (must match some area-name in screen-areas-summary)'',
        ``covered-by'': ``Name of the higher-level exploration-space or menu under which these elements will naturally be explored'',\newline

        ``reason-not-explored-independently'': ``Short explanation of why these elements do not need their own top-level exploration items (e.g., `will be covered when exploring the File menu and its submenus')''\newline

      \}\}
    ],\newline

    ``low-priority-areas'': [
    \newline
      \{\{
        ``area-name": ``Area or region that appears low-priority or mostly static'',\newline
        
        ``reason-low-priority'': ``Why this area is unlikely to require focused exploration (e.g., decorative logo, static background, non-interactive status bar)''\newline

      \}\}
    ]
  \}\}\newline

  \textbf{IMPORTANT:}\newline
  
  The area names you use in exploration-plan keys MUST be exactly the same as the area-name values you create in screen-areas-summary. Analyze the screenshot to determine what areas exist, then use those same names consistently in both places.\newline

  \textbf{CRITICAL GUIDELINES} for exploration-space: \newline

  1. If the element has visible text/label: Use that text exactly as it appears (e.g., ``File", ``Edit'', ``Opacity'', ``New Layer'')\newline

  2. If the element has NO visible text/label: You MUST provide BOTH:\newline

    - Shape description: What the icon/button looks like (e.g., ``dotted rectangle'', ``brush icon'', ``eyedropper shape'', ``circular button with gradient''')
    \newline
    \newline
    - Position description: Precise location within its area (e.g., ``first row second column'', ``top row leftmost'', ``second row third column'', ``bottom section left side'')
    - Example: ``the button in the left toolbox, first row second column, shaped like a dotted rectangle with dashed border''\newline

    - Example: ``the icon in the right dock, top section, third from left, shaped like a brush/paintbrush''\newline
    
    - Example: ``the slider in the left panel, middle section, labeled `Opacity' or appearing as a horizontal bar''\newline

  \textbf{Important guidelines:}
\newline

  - \textbf{OBSERVATION-BASED ANALYSIS:} Describe only what you can actually see in the screenshot. Do not infer functionality based on names or icons that might remind you of other software. \newline

  - \textbf{AREA IDENTIFICATION:} Analyze the screenshot to identify distinct UI areas. Name them based on:
  \newline
  \newline
    * Their location in the window (top, left, right, bottom, center)
    \newline
    
    * Their visual appearance (horizontal bar, vertical panel, canvas area, dock panel, toolbox)
    \newline
    
    * Their content (text menus, icon buttons, image display, etc.)\newline
    
    * Example naming patterns: ``Top horizontal bar with text labels'', ``Left vertical panel with icons'', ``Center canvas area'', ``Right side panel with tabs'' \newline
    
    * DO NOT use generic names like "Top Menu Bar" unless you actually see a menu bar at the top. Base names on what you observe in the screenshot. \newline
    
  - \textbf{AREA NAME CONSISTENCY:} \newline
  
  The area names you create in screen-areas-summary MUST be used as keys in exploration-plan. Each area-name in screen-areas-summary should have a corresponding key in exploration-plan with the exact same name. This creates the hierarchical connection between the summary and the plan. \newline
  
  - \textbf{HIERARCHICAL ORGANIZATION:} \newline

  Organize exploration-plan by the areas you identified. The structure should be: \{\{``Area Name 1'': [items], ``Area Name 2'': [items], ...\}\} where Area Names exactly match the area-name values from screen-areas-summary. The items listed in exploration-plan should be restricted to the HIGHEST-LEVEL menus/controls and major functional regions.
  \newline
  
  - \textbf{COMPREHENSIVE COVERAGE:}\newline
  
  Ensure that ALL interactive elements you can observe are accounted for EITHER as: \newline
  
    * top-level items in exploration-plan (for the highest-level menus/controls that should be explored directly), OR \newline
    
    * grouped entries in lower-hierarchy-elements (for buttons/controls that will be naturally explored when interacting with a higher-level item), OR \newline
    
    * entries in low-priority-areas (for areas that appear mostly static or low-value, with an explanation). \newline
    
  - \textbf{NO OVERLAP / MUTUAL EXCLUSIVITY:}\newline
  
  Do NOT place the same concrete interactive element or area in more than one of these structures. Every element/area must belong to exactly ONE of: exploration-plan, lower-hierarchy-elements, or low-priority-areas. Use the Stage 1 menu hierarchy summary to decide whether a region is best treated as a root, a lower-hierarchy group, or low priority. \newline
  
  - \textbf{POSITION DETAILS:} \newline
  
  For elements without text labels, ALWAYS include:\newline
  
    * Row/column position (e.g., ``first row'', ``second column'')\newline
    
    * Relative position within the area (e.g., ``top section'', ``middle'', ``bottom left'')\newline
    
    * Visual shape/icon description\newline
    
    * This helps the grounding model locate elements accurately\newline

  \textbf{CRITICAL GUIDELINES FOR instruction:}\newline
  
  - Each instruction should be a **single, short next-step action** the agent can take immediately based on the current observation (e.g., ``Click `File''', ``Open the Tools menu'', ``Click the zoom dropdown at the bottom right'').\newline
  
  - Avoid long multi-step or comprehensive phrases like ``fully explore'', ``systematically inspect every option'', or ``explore all submenus''. Those broader behaviors will be handled by higher-level controllers, not in this instruction field.\newline
  
  - The instruction should:\newline
  
    * Specify exactly ONE main interaction (e.g., click, open, hover, type a short string, toggle a control).\newline
    
    * Be grounded in visible UI elements and the Stage 1 hierarchy (e.g., refer to the exact label or clear visual description).\newline
    
  - Instructions must be concrete, concise, and observation-based, focused on the immediate next action rather than an entire exploration sequence.\newline

  - Base exploration on what hasn't been explored yet according to the history\newline
  
  - Focus on the application window only, ignore OS-level UI\newline
  
  - \textbf{REMEMBER:} You have NO prior knowledge about this software. Base everything on the screenshot observation. The area names should come from your analysis of the screenshot, not from prior knowledge or assumptions.\newline

  Provide your response in JSON format only.
\end{InitInteractPromptBox}

% ---------- Feedback: state classification ----------
\begin{StateClassifyPromptBox}
 You are a feedback model for GUI exploration. Please follow the instructions below to classify the exploration step.\newline

        The area the agent is actively exploring is:
        \textit{\{node-path\}}  \newline

        The recent instruction for the exploration step is: 
        \textit{\{instruction\}}

        The historical exploration action sequences (step-wise) in this DFS step are:
        \{history-actions-text\}\newline

        The stop reason for this step is:
        \textit{\{history-summary\}} \newline

        You are given TWO screenshots: \newline
        
        1) The original screenshot BEFORE these actions. \newline
        
        2) The new screenshot AFTER these actions.\newline

        First, carefully read the following FOUR SITUATIONS that describe typical exploration outcomes. \newline
        
        Your job is to decide which ONE situation best matches the current case. \newline

        \textbf{Situation 1 (non-terminal but reasonable)}:\newline
        
        - The instruction is completed and the actions are overall reasonable. \newline
        
        - However, this step does NOT correspond to a final unit function; more exploration is still needed from the current screen. \newline
        
        - \textbf{Typical cases:} \newline
        
          * The agent clicks a top-level menu (e.g., ``File'') and a menu opens, but no specific sub-function (like ``Save As'') has been fully explored yet.\newline
          
          * The agent opens a side panel or toolbar that reveals many options, but does not yet test any particular option in detail.\newline
          
        - \textbf{Potential Harder Cases:}\newline
        
          * While the screen shows no change, the click is the correct one, but only the sidebars to open by this action are already opened (that explains why nothing happens on the screen, but the click is still correct and should be counted as a reasonable action.)\newline

        \textbf{Situation 2 (terminal and reasonable)}:\newline
        
        - The instruction is completed and the actions are overall reasonable.\newline
        
        - This step DOES correspond to a completed unit function whose behavior can be summarized from the AFTER screenshot.\newline
        
        - \textbf{Typical cases:}\newline
        
          * The agent successfully applies a formatting change (e.g., makes selected text bold) and the visual change is clearly visible.\newline
          
          * The agent completes a dialog workflow (e.g., opens ``Save As'', chooses a name, and confirms), and the result is visible on screen.\newline
          
          * Usually, the button or menu that is clicked, its name shall be reasonable correspond to the screen change that you would be able to observe.\newline

        \textbf{Situation 3 (reasonable instruction, problematic actions):} \newline
        
        - The exploration instruction is reasonable for this UI.\newline
        
        - However, the concrete actions are NOT reasonable (e.g., wrong target, missing key steps, or obviously ineffective).\newline
        
        - This is often paired with a stop reason like ``max steps exceeded'' or no meaningful visual change.\newline
        
        \textbf{- Typical harder cases:}\newline
        
          * The instruction says ``Select the target text and apply bold to it'', but the agent failed to select the target text. Or it has no idea how to come up with a good way to perform this fine-grained action.\newline

        \textbf{Situation 4 (problematic instruction, roughly reasonable actions align with the instruction):} \newline
        
        - The action sequence itself is roughly reasonable for what the agent seems to be trying to do.\newline
        
        - However, the EXPLORATION INSTRUCTION is poorly posed for the current screen and the current target (underspecified, missing prerequisites, or misleading).\newline
        
        \textbf{- Typical harder cases:} \newline
        
          * The instruction says`` test/click the bold function'', but there is no text selected and the agent just clicks the Bold button once.\newline
          
          * The instruction says ``test/click the crop tool'', but no region is selected or adjusted after clicking the crop icon.\newline

        \textbf{YOUR TASK:}\newline
        
        1) Choose the most appropriate situation among 1, 2, 3, and 4.\newline
        
        2) Provide a brief natural-language reasoning string explaining why this situation fits best.\newline

       \textbf{ Output STRICTLY the following JSON:} \newline
        
        \{\{
          ``situation'': 1 or 2 or 3 or 4,
          ``reasoning'': ``brief explanation of why you chose this situation''
        \}\}
\end{StateClassifyPromptBox}

% ---------- Feedback: terminal state verification ----------
\begin{TerminalPromptBox}
You are a feedback model for GUI exploration. This may be ANY application (known or unknown). \newline

\textbf{Your job:} analyze screenshots and ground feedback to VISUAL OBSERVATIONS ONLY.\newline

\textbf{CONTEXT:}\newline

\textbf{Exploring:} \textit{\{node-path\}}\newline

\textbf{Instruction:} \{instruction\}\newline

\textbf{Actions:} \{history-actions-text\}\newline

\textbf{Stop reason:} \{history-summary\}\newline

\textbf{Classification:} \{reasoning\}\newline

\textbf{Screenshots:} BEFORE (state before actions) and AFTER (state after actions)\newline

\textbf{SITUATION 1:} Non-terminal but reasonable - more exploration needed from current state.\newline

\textbf{YOUR TASK:} Design NEXT exploration items that CONTINUE from this step.\newline

\textbf{REQUIREMENTS:}\newline

\textbf{1. **GROUND TO VISUALS** (Priority \#1):}\newline

   - Examine BEFORE/AFTER screenshots carefully\newline
   
   - Base ALL suggestions on what you SEE in images\newline
   
   - Identify what changed: new menus, panels, dialogs, buttons\newline
   
   - DO NOT fabricate features based on assumptions about this software type\newline
   
   - Use general UI patterns (buttons, menus, dialogs), not software-specific knowledge\newline

\textbf{2. **Describe Elements Visually**:}\newline

   - If unclear labels: use position + visual description\newline
   
   - Examples: ``second button from left with bold `B' icon'', ``third menu item in dropdown'', ``circular blue button at bottom-right''\newline
   
   - DO NOT invent names - describe what you actually see\newline

\textbf{3. **Sequential vs. Combined**:} \newline

   - SEPARATE items: Independent buttons/options → one item per element, include ALL visible ones\newline
   
   - COMBINED item: Multi-step workflow (e.g., dialog with fields+button) → one item with all steps\newline

\textbf{4. **Continuation Logic**:}\newline

   - Only suggest elements VISIBLE in AFTER screenshot\newline
   
   - Only suggest elements that appeared/became accessible from this step\newline
   
   - Menu opened → explore visible menu items\newline
   
   - Panel appeared → explore visible controls in it\newline
   
   - Dialog opened → explore visible options in it\newline
   
\textbf{5. **DO NOT**: }\newline

   - List all clickable elements\newline
   
   - Suggest elements not visible in screenshots\newline
   
   - Assume features that "should" exist\newline
   
   - Jump to unrelated areas\newline

\textbf{OUTPUT (JSON only):}\newline

        \{\{\newline
        
          ``next-exploration-items'': [\newline
          
            \{\{\newline
            
      ``area'': ``VISIBLE area description (e.g., `dropdown menu at top', `toolbar second row')'',\newline
      
      ``exploration-space'': ``Element description you can SEE (e.g., `third button with scissors icon', `Save As menu item')'',\newline
      
      ``instruction'': ``Clear instruction based on VISIBLE elements''\newline
      
    \}\}\newline
    
  ]\newline
  
\}\}\newline

\textbf{Requirements:}\newline

- Non-empty list\newline

- Each item describes VISUALLY CONFIRMED elements in AFTER screenshot\newline

- Grounded to what changed from this step\newline

- No fabrication\newline

\end{TerminalPromptBox}

% ---------- Feedback: error attribution (planning) ----------
\begin{PlanErrorPromptBox}
You are a feedback model for GUI exploration with deep knowledge of various desktop applications and their functionalities.\newline

\textbf{CONTEXT:} \newline

\textbf{CONTEXT:}\newline

\textbf{Exploring:} \textit{\{node-path\}}\newline

\textbf{Instruction:} \{instruction\}\newline

\textbf{Actions:} \{history-actions-text\}\newline

\textbf{Stop reason:} \{history-summary\}\newline

\textbf{Classification:} \{reasoning\}\newline

\textbf{Screenshots:} BEFORE (state before actions) and AFTER (state after actions)\newline

\textbf{SITUATION 2: Terminal and Reasonable Exploration}\newline

The instruction was successfully completed. This represents a COMPLETE unit function that should be documented.\newline

\textbf{YOUR TASK:}\newline

Create a HIGH-QUALITY function summary that accurately captures what was accomplished.\newline

\textbf{CRITICAL REQUIREMENTS:}\newline

\textbf{1. **Use Your Prior Knowledge**:}\newline

Apply your understanding of application features and typical user workflows to interpret what happened.\newline

\textbf{2. **Trajectory-Based Summary**:}\newline

   - Analyze the COMPLETE action trajectory (buttons clicked, menus accessed, etc.)\newline
   
   - Identify the EXACT functionality that was exercised\newline
   
   - Note the visual changes in the AFTER screenshot\newline

\textbf{3. **Summary Quality**:}\newline

   - Be SPECIFIC about what the function does (not just "clicked a button")\newline
   
   - Reference the UI elements involved (button names, menu paths, etc.)\newline
   
   - Describe the OBSERVABLE effect or outcome\newline
   
   - Keep it concise (1-2 sentences) but informative\newline

\textbf{4. **Examples of Good Summaries**:}\newline

   - ``Applied bold formatting to selected text by clicking the Bold button in the toolbar, resulting in the text appearing in bold weight.''\newline
   
   - ``Opened the Save As dialog from File menu and successfully saved the document with a new filename to the Desktop location.''\newline
   
   - ``Inserted a table with 3 rows and 2 columns using the Insert > Table menu option, which now appears in the document.''\newline

\textbf{5. **Examples of Poor Summaries**:}\newline

   - ``Clicked a button.'' (too vague)\newline
   
   - ``Did something with text.'' (not specific)\newline
   
   - ``Explored the menu.'' (doesn't describe the function)\newline

\textbf{UTPUT FORMAT (JSON):}\newline

\{\{
  ``function-summary'': ``One or two sentences clearly describing the unit function, its trigger (buttons/menus), and its observable effect''
\}\}\newline

Return ONLY valid JSON. The function-summary field is CRITICAL and must be high-quality.\newline

\end{PlanErrorPromptBox}

% ---------- Feedback: error attribution (action) ----------
\begin{ActionErrorPromptBox}
You are a feedback model for GUI exploration with expertise in action sequences and UI interaction patterns.

\textbf{CONTEXT:}\newline

\textbf{Exploring:} \textit{\{node-path\}}\newline

\textbf{Instruction:} \{instruction\}\newline

\textbf{Actions:} \{history-actions-text\}\newline

\textbf{Stop reason:} \{history-summary\}\newline

\textbf{Classification:} \{reasoning\}\newline

\textbf{Screenshots:} BEFORE (state before actions) and AFTER (state after actions)\newline

\textbf{SITUATION 3: Reasonable Instruction, Problematic Actions}\newline

The exploration instruction is valid, but the ACTION SEQUENCE has issues (wrong targets, missing steps, ineffective execution).\newline

\textit{\{fine-grained-actions-text\}}\newline

\textbf{YOUR TASK:}\newline

Provide ACTIONABLE feedback to fix the action sequence. If the task requires fine-grained actions, you MUST select the appropriate action from the list above and provide its specific instructions.\newline

\textbf{CRITICAL REQUIREMENTS:}\newline

\textbf{1. **Use Your Prior Knowledge**:} Apply your understanding of typical UI interaction patterns to diagnose what went wrong.\newline

\textbf{2. **Careful Analysis**:}\newline

   - Compare what SHOULD have happened vs. what DID happen\newline
   
   - Identify specific problems: wrong element grounded, missing prerequisites, incorrect order, etc.\newline
   
   - Determine if a fine-grained action is needed\newline

\textbf{3. **Fine-Grained Action Selection**:}\newline

   - If the task requires a fine-grained action, you MUST:\newline
   
     * Choose the MOST APPROPRIATE action from the available fine-grained actions list above\newline
     
     * Include the EXACT action-instruction steps from that action in your feedback\newline
     
     * Include the action-primitive code if helpful for understanding\newline
     
   - DO NOT suggest retrying without providing the fine-grained action template\newline

\textbf{4. **Feedback Structure When Fine-Grained Action Required**:}\newline

   a) **Problem Diagnosis**: What went wrong and why\newline
   
   b) **Selected Fine-Grained Action**: Name of the chosen action (e.g., ``FineGrainedActions.select-text-span'')\newline
   
   c) **Action Instructions**: Copy the exact action-instruction steps from the selected action\newline
   
   d) **Action Primitives**: Copy the exact action-primitive code from the selected action\newline
   
   e) **FINE-GRAINED-REQUIRED: Yes** - Mark explicitly\newline

\textbf{5. **Feedback Structure When No Fine-Grained Action Needed**:}\newline

   a) **Problem Diagnosis**: What went wrong\newline
   
   b) **Suggested Fix**: Concrete correction steps using standard actions\newline
   
   c) **FINE-GRAINED-REQUIRED: No**\newline

\textbf{OUTPUT FORMAT (JSON):}\newline

\{\{\newline

  ``feedback'': ``Your detailed feedback following the structure above'',\newline
  
  ``fine-grained-required'': true or false,\newline
  
  ``selected-fine-grained-action'': ``Action name if applicable, otherwise omit this field''\newline
  
\}\}\newline

\textbf{Example with fine-grained action:}\newline

\{\{\newline

  ``feedback'': The agent is instructed to select a target text span, however, due to the fine-grained nature of the action, the action model failed to produce the action successfully.\newline
  
  Selected Fine-Grained Action:\newline
  
  FineGrainedActions.select-text-span Action Instructions:\newline
  
  1. Move the cursor to the start of the target text span in the editor.\newline
  
  2. Press and hold the left mouse button.\newline
  
  3. Drag the cursor to the end of the target text span.\newline
  
  4. Release the left mouse button to finalize the selection.\newline
  
  Action Primitives:\newline
  import pyautogui; pyautogui.moveTo(START-X, START-Y, duration=0.1);\newline
  
  pyautogui.mouseDown(); pyautogui.moveTo(END-X, END-Y, duration=DURATION);\newline
  
  pyautogui.mouseUp(); FINE-GRAINED-REQUIRED: Yes'',\newline

  ``fine-grained-required'': true,\newline
  
  ``selected-fine-grained-action'': ``FineGrainedActions.select-text-span''\newline
  
\}\}\newline

\textbf{Example without fine-grained action:}\newline

\{\{\newline

  ``feedback'': ``Problem: The agent clicked the wrong button.\newline
  
  Suggested Fix: Click the correct `Save' button instead of `Cancel'.\newline
  
  FINE-GRAINED-REQUIRED: No'',\newline
  
  ``fine-grained-required'': false\newline
  
\}\}\newline

Return ONLY valid JSON. If fine-grained action is required, you MUST select one from the available list and provide its instructions.
\end{ActionErrorPromptBox}

\end{document}